\documentclass[10pt,journal,compsoc]{IEEEtran}

\ifCLASSOPTIONcompsoc
  \usepackage[nocompress]{cite}
\else
  \usepackage{cite}
\fi

\usepackage{xcolor}
\definecolor{codegreen}{rgb}{0,0.6,0}
\definecolor{codegray}{rgb}{0.5,0.5,0.5}
\definecolor{codepurple}{rgb}{0.58,0,0.82}
\definecolor{backcolour}{rgb}{0.95,0.95,0.92}
\usepackage{listings}
\lstdefinestyle{mystyle}{
  backgroundcolor=\color{backcolour}, commentstyle=\color{codegreen},
  keywordstyle=\color{magenta},
  numberstyle=\tiny\color{codegray},
  stringstyle=\color{codepurple},
  basicstyle=\ttfamily\footnotesize,
  breakatwhitespace=false,         
  breaklines=true,                 
  captionpos=b,                    
  keepspaces=true,                 
  numbers=left,                    
  numbersep=5pt,                  
  showspaces=false,                
  showstringspaces=false,
  showtabs=false,                  
  tabsize=2,
  columns=fullflexible
}
\usepackage{graphicx}
\usepackage{url}                 
\usepackage[colorlinks,citecolor=red,urlcolor=blue,bookmarks=false,hypertexnames=true, pagebackref=true]{hyperref}            
\usepackage{booktabs}            
\usepackage{multirow}            
\usepackage{amsmath,amsfonts,amssymb}
\usepackage{xcolor}
\usepackage{algorithm}
\usepackage{algpseudocode}
\usepackage{algorithmicx}
\usepackage{import}
\usepackage{etoolbox}

\DeclareMathOperator*{\argmax}{arg\,max}

\newcommand{\cut}[1]{}

\usepackage{bm} 
\usepackage{subfig}
\usepackage{float}  
\usepackage{multirow}
\usepackage{xcolor,colortbl} 
\usepackage{color}  
\usepackage{amsfonts}  
\usepackage{mathtools}
\usepackage[capitalise, nameinlink]{cleveref} 
\newtheorem{theorem}{Theorem}

\usepackage{cite} 

\usepackage{grstyle}  

\usepackage{selectp} 

\providecommand{\citep}{\cite} 
\providecommand{\citet}{\cite}
\newtheorem{lemma}{Lemma}

\newtheorem{definition}{Definition}[section]

\newcommand{\overlengthPAMI}[1]{}

\providecommand{\colReb}{black}  
\providecommand{\rebuttal}[1]{{#1}}

\usepackage{subfiles} 

\usepackage{amsmath,amsfonts,bm}

\def\eqref#1{equation~\ref{#1}}

\def\1{\bm{1}}

\def\vc{{\bm{c}}}

\def\vv{{\bm{v}}}
\def\vw{{\bm{w}}}

\def\vz{{\bm{z}}}

\DeclareMathAlphabet{\mathsfit}{\encodingdefault}{\sfdefault}{m}{sl}
\SetMathAlphabet{\mathsfit}{bold}{\encodingdefault}{\sfdefault}{bx}{n}

\newcommand{\R}{\mathbb{R}}

\begin{document}
\title{Hadamard product in deep learning: \\ Introduction, Advances and Challenges}

\ifCLASSOPTIONcompsoc  
\author{%
Grigorios G Chrysos, \quad Yongtao Wu, \quad Razvan Pascanu, \quad Philip Torr, \quad Volkan Cevher\vspace{2mm} 
\IEEEcompsocitemizethanks{\IEEEcompsocthanksitem GC is with the University of Wisconsin-Madison.
\IEEEcompsocthanksitem YW, VC are with the Department of Electrical Engineering, Ecole Polytechnique Federale de Lausanne (EPFL), Switzerland.
\IEEEcompsocthanksitem RP is with DeepMind.
\IEEEcompsocthanksitem PT is with the Department of Engineering Science, University of Oxford.%

Corresponding author's e-mail: chrysos@wisc.edu.}
\thanks{}
}

\IEEEtitleabstractindextext{%
\begin{abstract}

    \rebuttal{While convolution and self-attention mechanisms have dominated architectural design in deep learning, this survey examines a fundamental yet understudied primitive: the \emph{Hadamard product}. Despite its widespread implementation across various applications, the Hadamard product has not been systematically analyzed as a core architectural primitive. We present the first comprehensive taxonomy of its applications in deep learning, identifying four principal domains: higher-order correlation, multimodal data fusion, dynamic representation modulation, and efficient pairwise operations. The Hadamard product's ability to model nonlinear interactions with linear computational complexity makes it particularly valuable for resource-constrained deployments and edge computing scenarios. We demonstrate its natural applicability in multimodal fusion tasks, such as visual question answering, and its effectiveness in representation masking for applications including image inpainting and pruning. This systematic review not only consolidates existing knowledge about the Hadamard product's role in deep learning architectures but also establishes a foundation for future architectural innovations. Our analysis reveals the Hadamard product as a versatile primitive that offers compelling trade-offs between computational efficiency and representational power, positioning it as a crucial component in the deep learning toolkit.}

\end{abstract}

}


    \maketitle
    \IEEEdisplaynontitleabstractindextext
    
    %
    \IEEEpeerreviewmaketitle

\section{Introduction}
\label{sec:elementwise_pr_intro}

\IEEEPARstart{D}{eep} neural networks (DNNs) have demonstrated unparalleled success in learning complicated concepts. Diverse tasks (such as image classification~\cite{hu2018squeeze}, image generation~\cite{karras2018style, park2019semantic}, image inpainting~\cite{pathak2016context, yu2018generative}, language understanding~\cite{dauphin2017language, devlin2019bert}, sequence modeling~\cite{hochreiter1997long, cho2014properties}, sequence transduction~\cite{vaswani2017attention} and multimodal learning~\cite{antol2015vqa, fukui2016multimodal}) are nowadays predominantly modelled with DNNs. One of the pillars behind the success of DNNs (and the focus of this work) is the choice of the architectural components, and more specifically the type of correlations captured.

Two of the most popular architectural components used across domains and tasks are the convolutional layer and the self-attention~\cite{vaswani2017attention}. A convolutional layer represents an element as a weighted sum of the neighboring input elements. Concretely, given a datapoint $\minvar$ (e.g., a sequence of text tokens or an image), a convolutional layer expresses each output as $\sum_{i, j} w_{i, j} \invar_{i, j}$ for (learnable) weights $\bm{W}$. A convolutional layer captures linear correlations between the input elements, which limits the contextual information. On the contrary, the self-attention (SA) represents nonlinear interactions through a matrix multiplication of the input elements, i.e., products of the form $(\bm{W}_1 \minvar) \cdot (\bm{W}_2 \minvar)$ for (learnable) weights $\bm{W}_1, \bm{W}_2$. However, SA suffers from a quadratic computational cost over the number of input elements~\cite{keles2023computational}, which results in a significant computational overhead for deep networks. Instead, the Hadamard product captures \emph{nonlinear interactions with a linear computational cost}. The Hadamard product is a pairwise operator that accepts two input tensors and outputs their element-by-element product in each position. Representing nonlinear interactions with a linear computational cost is critical in the era of deep networks, e.g., in order to use them on edge devices~\cite{babiloni2021poly, hua2022transformer} or especially for large language models and foundational models~\cite{bommasani2021opportunities} that are computationally intensive.\looseness-1 


\begin{table*}[ht]
            \centering
            \caption{\rebuttal{Recent works using the Hadamard product. Beyond the seminal works that Hadamard product has been used, the Hadamard product exists in new applications, e.g., in recent language models or parameter-efficient fine-tuning.}}
            
            { \color{\colReb} 
            \begin{tabular}{c@{\hspace{0.65cm}} p{4.4cm}@{\hspace{0.45cm}} c@{\hspace{0.45cm}} m{3.5cm}}
                Category                                    & Task                              & Method                                            & Publication\\\hline
                \multirow{4}{*}{High-order interactions}    
                                                            & Video action recognition      & \cite{wasim2023video}     & ICCV'23\\
                                                            & Image classification         & MONet~\cite{chen2024multilinear}                     & ICLR'24\\
                                                            & Masked language modeling         & M2~\cite{fu2023monarch}                            & NeurIPS'23\\
                                                            & Classification/segmentation              & RMT~\cite{fan2024rmt}                     & CVPR'24\\
                                                            \hline
                \multirow{3}{*}{Multimodal fusion}         
                                                            & Video Question Answering   & \cite{zou2024language}                               & CVPR'24\\
                                                            & Image generation                      & MultiDiffusion~\cite{bar2023multidiffusion}      & ICML'23 \\
                                                            & Image generation                      & DenseDiffusion~\cite{kim2023dense}      & ICCV'23 
                                                            \\\hline
                \multirow{10}{*}{Adaptive modulation}        
                                                            & Language modeling                 & xLSTM~\cite{hochreiter1997long}                    & NeurIPS'24\\
                                                            & Classification/segmentation              & V-RWKV~\cite{duan2024vision}                         & ICLR'25\\
                                                            & Pruning                  & PX~\cite{iurada2024finding}                          & CVPR'24\\
                                                            & 3D texture generation                           & \cite{cao2023texfusion}       & ICCV'23 \\
                                                            & Conditional image generation                  & LayoutFlow~\cite{guerreiro2024layoutflow}                          & ECCV'24\\
                                                             & Language modeling                  & \cite{yang2024parallelizing}                           & NeurIPS'24\\
                                                             & Language modeling                 & Mamba~\cite{gu2024mamba}  & COLM'24\\ 
                                                             & Language modeling                 & Mamba2~\cite{dao2024transformers}  & ICML'24\\ 
                                                            & Language modeling                 & HGRN2~\cite{qin2024hgrn}  & COLM'24\\ 
                                                            & Language modeling                 & 
                                                            GLA~\cite{yang2024gated}
                                                            & ICML'24\\
                                                            \hline
                                                            
                \multirow{6}{*}{Efficient operator}         
                                                            & Image classification              & Poly-SA~\cite{babiloni2023linear}                   & PAMI'23\\
                                                            & Image classification              & StarNet~\cite{ma2024rewrite}                   & CVPR'24\\
                                                            & Masked language modeling          & FLASH~\cite{hua2022transformer}                   & ICML'22\\
                                                             &     Parameter-efficient fine-tuning     &  LoHa~\cite{hyeonwoo2022fedpara}                  &ICLR'22 \\
                                                            &     Parameter-efficient fine-tuning     &   HiRA~\cite{anonymous2025hira}                 & ICLR'25\\
                                                          &     Parameter-efficient fine-tuning     & MLAE~\cite{wang2024mlae}                   & Arxiv, May'24 \\
                                                             \hline
                
            \end{tabular}
            }
            \label{tab:hadamard_product_indicative_work_table}
\end{table*}
\vspace{2mm}

Capturing nonlinear interactions is especially important when we have different modalities. 
As humans, we perceive the world through various sensory input, e.g., visual and auditory modalities, that offer complementary information, while they share a common underlying structure. Enabling neural networks to process and understand multimodal information is critical for important applications, such as medical analysis. Clearly, data from different modalities might vary widely in volume, making processing more difficult. A core part of processing multimodal data is the fusion of those volumes. A way to fuse information is through naive concatenation, however this expresses only linear interactions~\cite{chrysos2021conditional}. Richer, nonlinear interactions can be captured with tensor decompositions, such as the popular CP decomposition. If we use the CP decomposition, the Hadamard product emerges naturally through standard properties (cf. the mixed product~\cite{kolda2009tensor} or \cref{lemma:polygan_lemma_hadamard_kr2}). Naturally, the Hadamard product has been widely applied to multimodal fusion tasks, such as in visual question answering~\cite{antol2015vqa} or multimodal medical analysis~\cite{shi2017multimodal}.\looseness-1 

Beyond the linear computational cost and the multimodal fusion, we argue that the Hadamard product lies at the heart of deep learning (DL) with far-reaching applications.  
In this work, our aim is to contextualize those applications of the Hadamard product and to frame it as a core operator in deep learning. Concretely, our analysis focuses on four categories where the Hadamard product has been instrumental. The first category revolves around representing high-order correlations, e.g., as in StyleGAN~\cite{karras2018style} or \sne~\cite{hu2018squeeze}. The second category develops the fusion of multimodal data, where different communities have independently used the Hadamard product as a fusion scheme. Then, we present the use of Hadamard product for dynamic filtering of features, e.g., in case there are corrupted parts of the image in masking~\cite{pathak2016context, yu2018generative}, and dynamic modulation of the weights, e.g., in Dropout to avoid overfitting or in LSTM to avoid vanishing/exploding gradients~\cite{hochreiter1997long}. \rebuttal{Masking is also utilized in language modeling, e.g., with linear attention models~\cite{katharopoulos2020transformers}, to enable efficient training on text sequences. This is critical for more computationally efficient training of models predicting the next output token.} The last category we revisit the computational efficiency of the Hadamard product, e.g., for reducing the complexity of standard self-attention blocks. 
By linking these applications of the Hadamard product and presenting them in a unified perspective, we hope to provide a more comprehensive understanding of this important operator in deep learning. This could also help to connect seemingly unrelated components, such as the inductive bias of polynomial networks~\cite{wu2022extrapolation} and the inductive bias of LSTM. In \cref{tab:hadamard_product_indicative_work_table} \rebuttal{we review some recent works in each category, with additional, significant works being mentioned in the respective sections, e.g., \cref{tab:hadamard_product_older_important_works_multimodal_fusion,tab:hadamard_product_older_important_works_adaptive_modulation}}.  \rebuttal{All in all, the contributions of this survey are the following:}

\begin{itemize}
    \item \rebuttal{This work presents the first taxonomy of the Hadamard product, establishing significant connections among different categories, such as multimodal fusion and high-order interactions.}

    \item \rebuttal{Within each category, we explore how various concepts influence the application of the Hadamard product, for example, masking in adaptive modulation.}

    \item \rebuttal{We highlight the key theoretical properties of the Hadamard product that have been developed in isolation. We discuss the impact of these properties, such as the effect of spectral bias on high-order interactions.}

    \item \rebuttal{We identify important open problems in \cref{sec:elementwise_pr_discussion}, which are expected to stimulate future research on this topic.}
\end{itemize}

\begin{figure*}[!tbh]
    \centering
    \subfloat[\rebuttal{\textbf{High-order interactions:}} Regular NN (top) and how to include higher-order interactions (bottom)]{\includegraphics[width=0.3\textwidth]{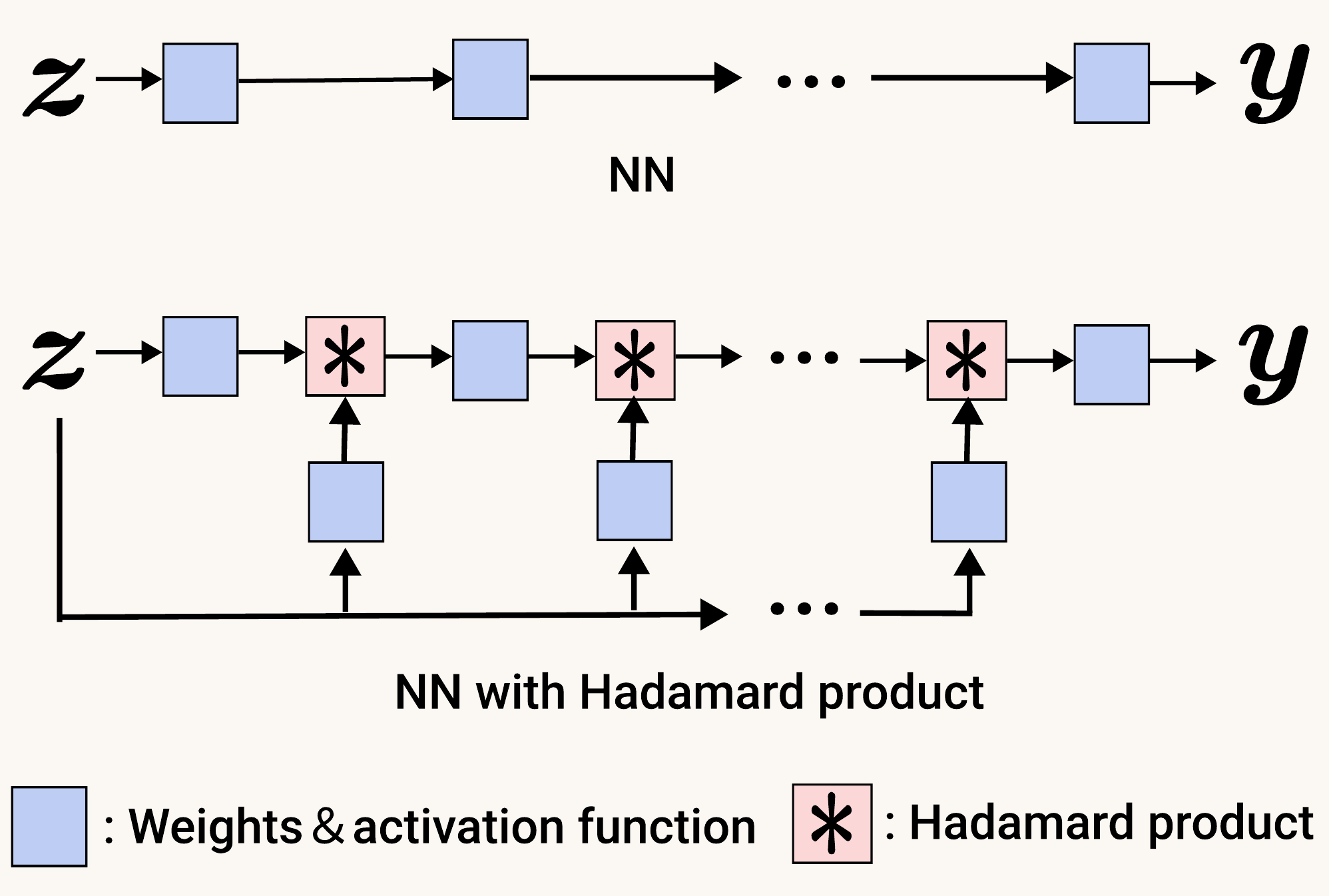}}
    \qquad
    \subfloat[\rebuttal{\textbf{Multimodal Fusion}}]{\includegraphics[width=0.3\textwidth]{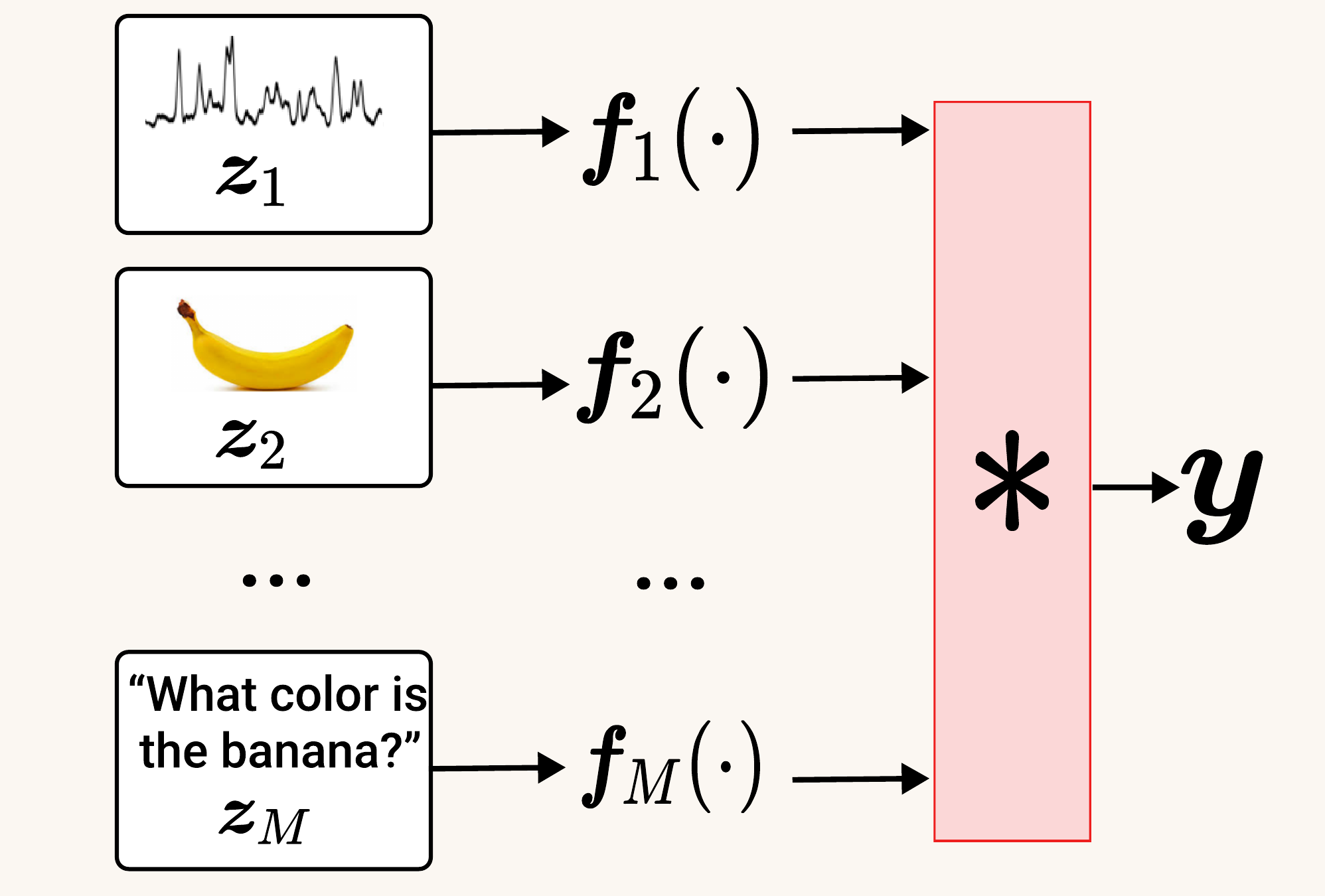}}
    \qquad
    \subfloat[\rebuttal{\textbf{Adaptive modulation:} Casual language modeling with linear attention}]{\includegraphics[width=0.3\textwidth]{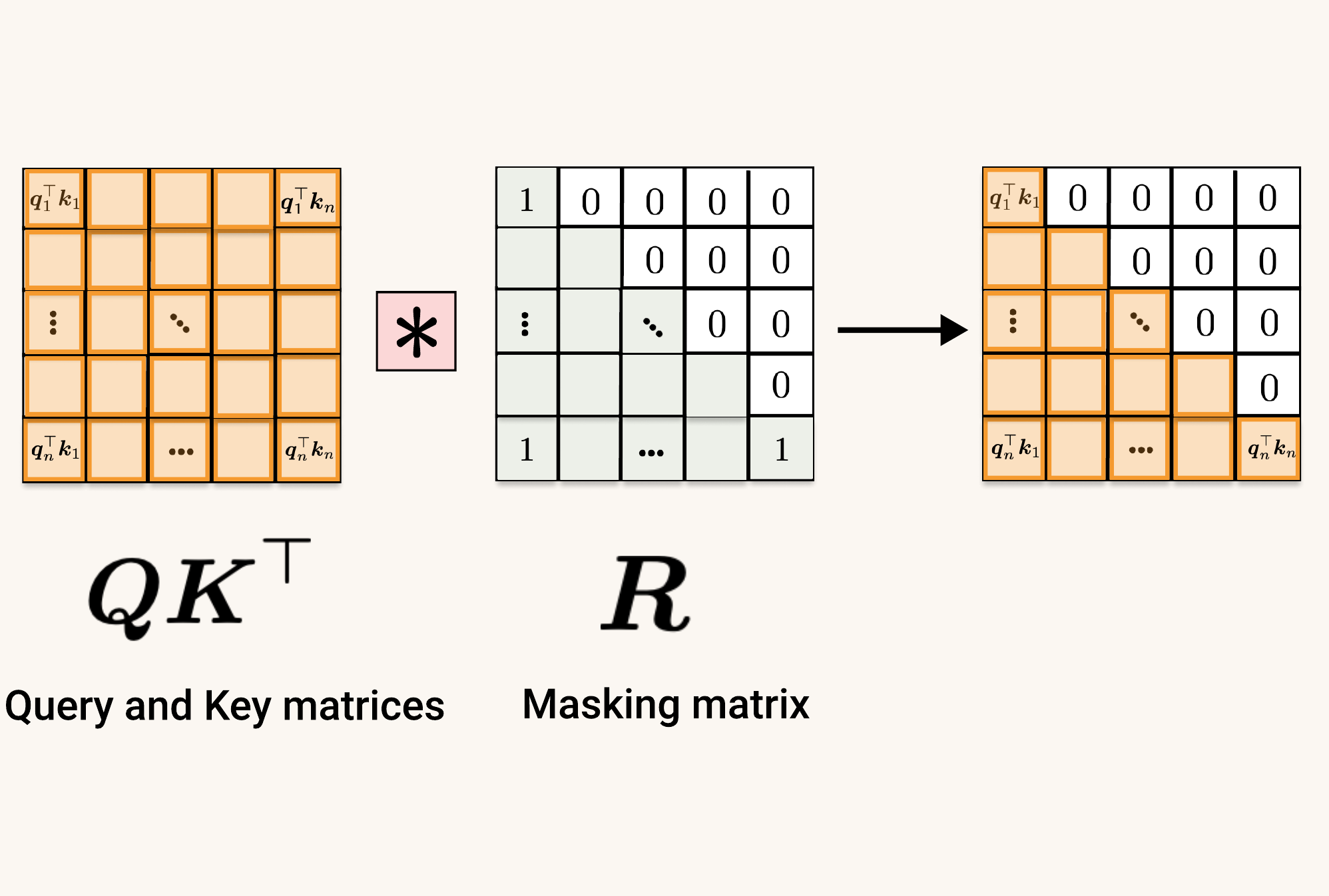}}
    \vskip\baselineskip
    \subfloat[\rebuttal{\textbf{Adaptive modulation:} Hard masking (top) and soft masking (bottom)}] {\includegraphics[width=0.3\textwidth]{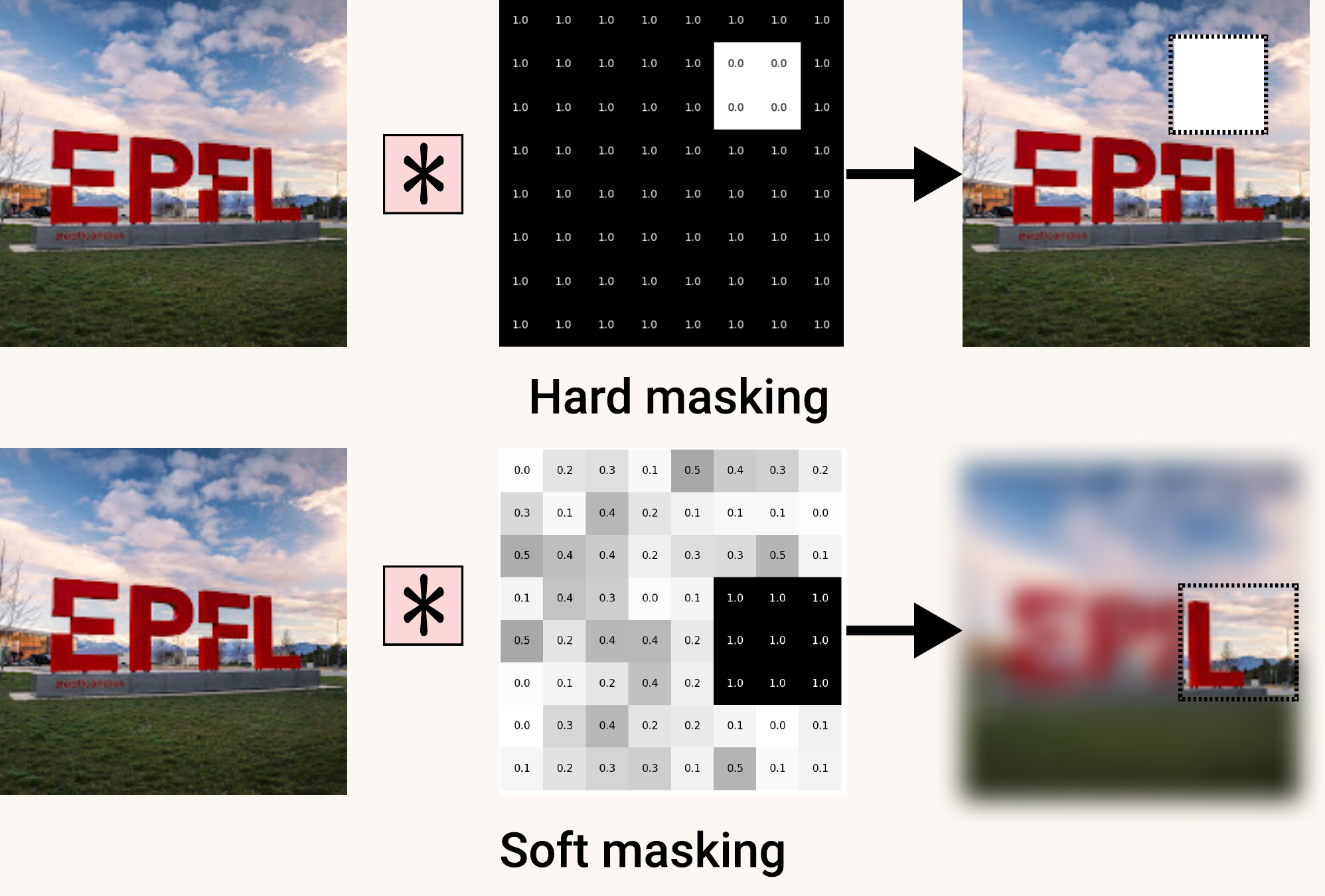}}
    \qquad
    \subfloat[\rebuttal{\textbf{Efficient operator}}]{\includegraphics[width=0.3\textwidth]{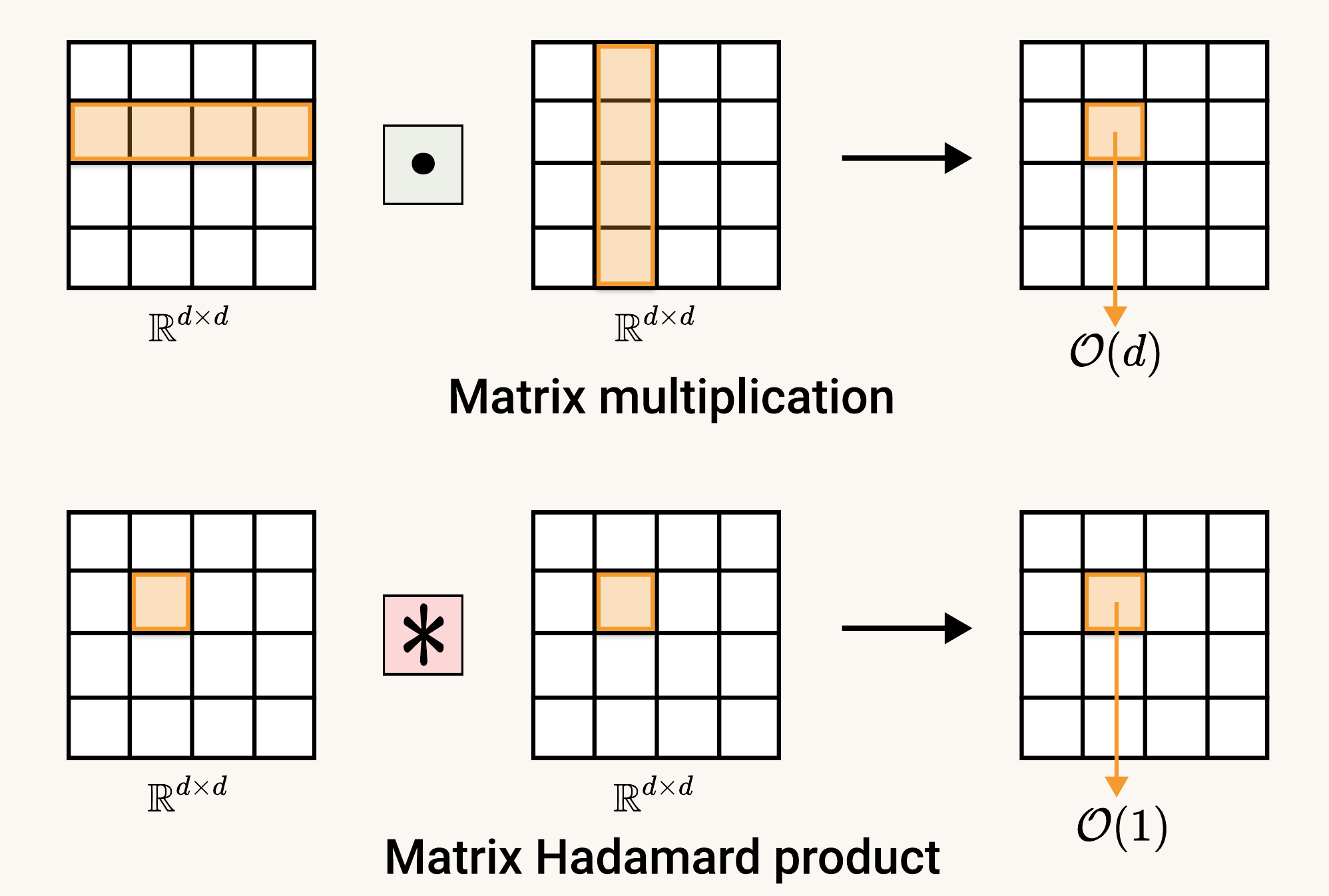}}
    \qquad
    \subfloat[\rebuttal{\textbf{Adaptive modulation and efficient operator:} Weight pruning}]{\includegraphics[width=0.3\textwidth]{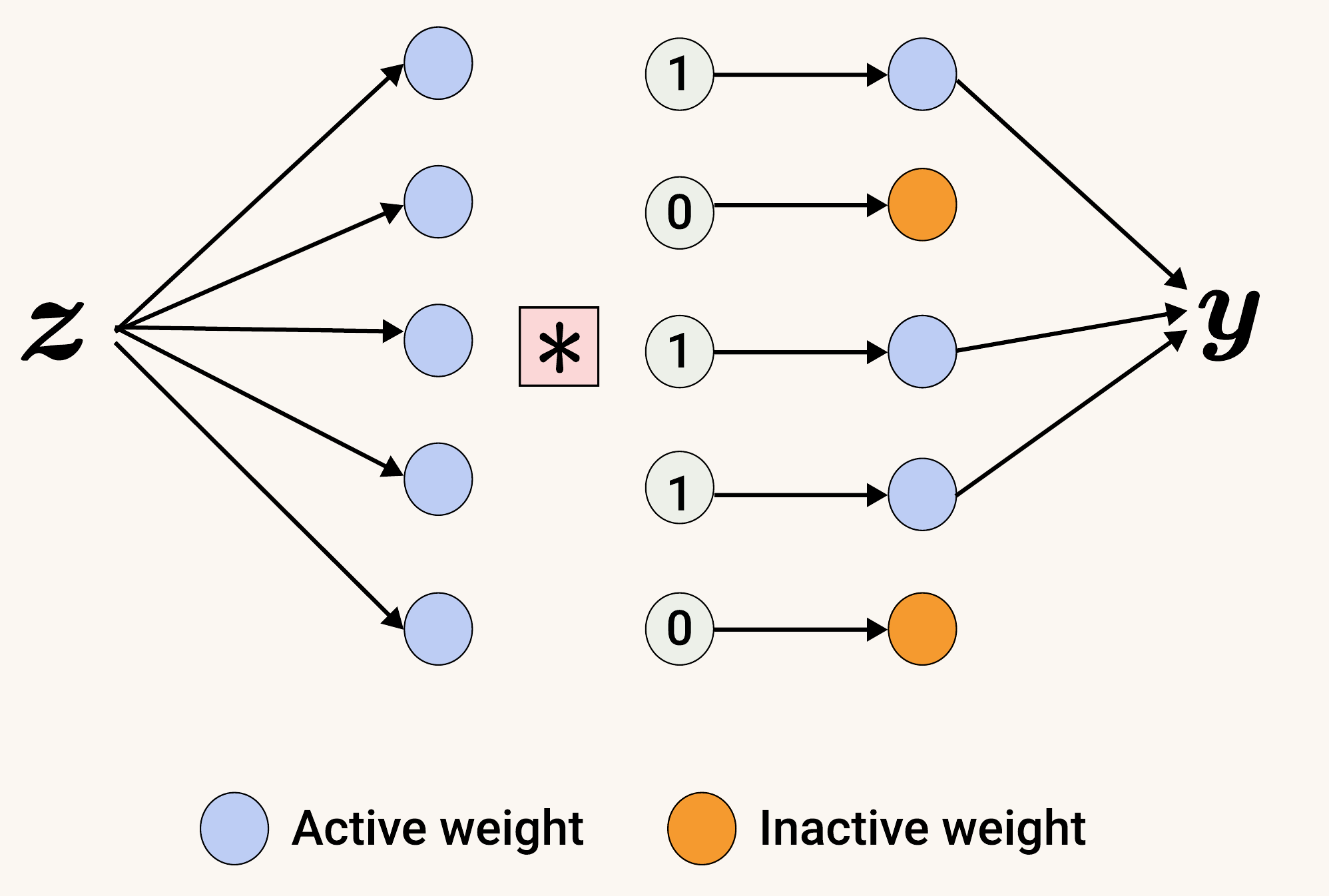}}
    \caption{
    \rebuttal{Six} core areas where the Hadamard product has been widely used in deep learning era. (a) High-order correlations between the input elements are captured. Those correlations can augment the linear interactions of the typical layers, e.g., dense or convolutional layers. (b) As humans we generally perceive the world through different senses, which often offer complementary information. Similarly, machine learning (ML) models can extract complementary information from different sources and then meld them together to make an informed decision. 
    \rebuttal{(c) During the pre-training of language modeling, we mask attention to the key of future tokens for each query so that the model does not use information from the next token when predicting the next token. }
    (d) Hard masking and soft masking via Hadamard product for the image in the input space.
    (e) The Hadamard product has recently been used as an alternative operator to the matrix multiplication, e.g., in order to accelerate the popular Self-Attention.
    \rebuttal{(f) Weight pruning can be viewed as applying a Hadamard product to the original weights, effectively zeroing out certain parameters. Among those core areas, we identify four parent categories and links between them (e.g. weight pruning). To our knowledge, this taxonomy is novel and allows us to establish concrete connections between seemingly disparate works within the same category, such as masking for inpainting and causal language modeling. To facilitate further research, we have also compiled the diverse open-source links in \cref{tab:hadamard_product_indicative_author_implementations}.}
    }
    \label{fig:elementwise_pr_motivational_fig}
\end{figure*}

\gnote{Notes for improvement of the figure, grigoris: 1) In (d) to mention: output of one element: O(n) in the case of mm, while O(1) in HP.}

\textbf{Brief history}: The term Hadamard product was originally termed in \cite{paul1948finite}. The influential book of \cite{styan1973hadamard} argues that the most popular references attribute the first use of this product in the manuscripts of Jacques Hadamard~\cite{hadamard1903leccons, hadamard1893resolution}. However, more recent books~\cite{johnson1990matrix} argue that this element-wise product had actually emerged at least few years earlier in Moutard's work. 
The works of Hadamard and Schur~\cite{schur1911bemerkungen} are among the first to (re-)invent the element-wise product. 
A by-product of the numerous re-inventions of this product is that there is no universally accepted symbol for the Hadamard product~\cite{johnson1990matrix}. In this work, we will interchangeably use the term Hadamard and element-wise product, while we denote this product with the symbol `$*$'.

\textbf{Relationship to other surveys}: \rebuttal{The rise in deep learning components and tasks has led to numerous surveys, covering areas like attention networks~\cite{han2022survey}, recurrent networks~\cite{yu2019review}, Dropout~\cite{labach2019survey}, tensors~\cite{kolda2009tensor, sidiropoulos2017tensor, panagakis2021tensor}, and multimodal learning~\cite{dalla2015challenges}. While the Hadamard product is related to some of these topics, such as Dropout and multimodal fusion, it is not the primary focus of these surveys. Our work fills this gap by explicitly addressing the Hadamard product and its wide range of applications, complementing existing surveys.}

\textbf{Paper outline}: 
We cover the first application of the Hadamard product in deep learning in \cref{sec:elementwise_pr_polynomial_nets}, while we provide a verbose taxonomy in \cref{fig:elementwise_pr_figure_taxonomy_visual}. Concretely, the Hadamard product is used there for capturing high-order correlations among the input elements, e.g., expressing a synthesized image as high-order correlations between the elements of the noise vector. This is a topic that has become relevant the last few years following the success of StyleGAN and other similar works. Another major area of research in the Hadamard product is the feature fusion, especially in the context of multimodal learning. We cover various feature fusion schemes using the Hadamard product in \cref{sec:elementwise_pr_feature_fusion}. A third area of development of networks with the Hadamard product is in masking, e.g., as in inpainting or Dropout. The idea is to modulate the input/features, such that the network can (exclusively) focus on certain areas of the representation. This idea has also been explored in the form of gating, which selects dynamically the critical features. Both masking and the gating mechanisms are detailed in \cref{sec:elementwise_pr_gating}. Furthermore, the Hadamard product has been featured recently as a more (computationally) efficient operator over the more costly matrix multiplication (e.g. in the context of self-attention) as covered in \cref{sec:elementwise_pr_efficient_op}. \rebuttal{Six popular applications of the four parent categories are visually depicted in \cref{fig:elementwise_pr_motivational_fig}.} Beyond the technical and computational aspects covered, the Hadamard product results in a number of interesting theoretical properties, which are studied in \cref{sec:elementwise_pr_theoretical_properties}. Lastly, we conduct a thorough discussion on the benefits of the Hadamard product, as well as highlight existing limitations and identify interesting avenues for future research on the topic in \cref{sec:elementwise_pr_discussion}.

\rebuttal{\textbf{Notation}: Matrices (vectors) are denoted by uppercase (lowercase) boldface letters e.g., $\bm{X}$, ($\bm{x}$). Tensors are denoted with boldface, calligraphic letters, e.g.,  $\bmcal{X}$. Detailed notation, along with core mathematical properties appear in \cref{sec:elementwise_pr_preliminaries}. In this work, the input to the network is denoted with (vector) $\binvar$ or (matrix) $\minvar$.}

\begin{definition}[Hadamard product]
    The \emph{Hadamard product} between two tensors $\bmcal{X}, \bmcal{Y}$ is denoted as $(\bmcal{X} \hadp \bmcal{Y}) \in \realnum^{I_1 \times I_2 \times \cdots \times I_M}$ and is defined as 
 $$(\bmcal{X} \hadp \bmcal{Y})_{i_{1}, i_{2}, \ldots, i_{M}}  \doteq (\bmcal{X})_{i_{1}, i_{2}, \ldots, i_{M}}\cdot (\bmcal{Y})_{i_{1}, i_{2}, \ldots, i_{M}}\;.$$
\end{definition}

\rebuttal{Notice that the Hadamard product between vectors and matrices are special cases of this definition.} 
Although the Hadamard product requires that the two tensors share the same dimensions, each $I_i$ can be different from $I_j$ for $i \neq j$. Further information on tensors, tensor decompositions and products can be found in the classic review papers of \cite{kolda2009tensor, sidiropoulos2017tensor}, while \cite{panagakis2021tensor} provides a recent overview of tensor in deep learning.

\section{High-order interactions} 
\label{sec:elementwise_pr_polynomial_nets}

Beyond the linear interactions used in neural networks, polynomial expansions have been explored for approximating signals with unknown analytic expressions. The Stone-Weierstrass theorem~\cite{stone1948generalized} guarantees that any smooth function can be approximated by a polynomial expansion, with its multivariate extension covered in standard textbooks, e.g., in \cite{nikol2013analysis} (pg 19). In this section, we analyze how such a polynomial expansion can be used to augment the linear interactions of the typical layers used in DNNs. Let $\binvar \in \realnum^d$ denote the input, e.g., an image patch, and let $\boutvar \in \realnum^o$ be the target, e.g., a class label. A polynomial expansion of the input $\binvar$ can be considered for approximating the target $\boutvar$. That is, a vector-valued function ${f}(\bm{z}): \realnum^{d} \to \realnum^{o}$ expresses the multivariate polynomial expansion of degree $N$:  \gnote{TODO: a) remove the function G, b) think whether the input/output can be defined in the notation if possible.}
\begin{equation}\label{eq:prodpoly_poly_general_eq}
    \boutvar = {f}(\binvar) = \sum_{n=1}^N \bigg(\bmcal{W}^{[n]} \prod_{j=2}^{n+1} \times_{j} \binvar\bigg) + \bm{\beta}\;,
\end{equation}
where $\bm{\beta} \in \realnum^o$ and $\big\{\bmcal{W}^{[n]} \in  \realnum^{o\times \prod_{m=1}^{n}\times_m d}\big\}_{n=1}^N$ are the learnable parameters. The form of \cref{eq:prodpoly_poly_general_eq} can approximate smooth functions (for large $N$). The number of parameters required to accommodate high-order interactions of the input increases exponentially with the desired degree of the polynomial which is impractical for high-dimensional data. Therefore, a form of parameter reduction is applied to each work to make polynomial expansions practical. 

Group Method of Data Handling (GMDH)~\cite{ivakhnenko1971polynomial} was one of the first works to learn quadratic polynomials. To reduce the parameteres, GMDH defines a predetermined set of interactions between the input elements. The method was later extended in high-degree polynomials~\cite{oh2003polynomial}. Contrary to the aforementioned predefined interactions, quadratic~\cite{hinton1983optimal, psaltis1986nonlinear} and high-degree expansions~\cite{sejnowski1986higher, giles1987learning} have been learned in a data-driven fashion. In particular, pi-sigma networks~\cite{shin1991pi} use a single multiplicative unit over the input elements. That is, given an input $\binvar \in \realnum^d$ the output $\boutvar \in \realnum^o$ in pi-sigma networks can be expressed as $\boutvar = \sigma(\bighp_{j=1}^J \bm{h}_j)$, with $\sigma$ an element-wise activation function and $J$ the order of the interactions. The vector $\bm{h}_j$ represents the $j^{\text{th}}$ column-vector of the hidden representation $\bm{H} = \bmcal{W}\times_3 \binvar + \bm{B}$. The parameters $\bmcal{W} \in \realnum^{o\times J \times d}, \bm{B} \in \realnum^{o\times J}$ are learnable. 

More recently, there are a number of works that use polynomial expansions in ML. Various quadratic polynomials of the form $f_1(\binvar) + f_2(\binvar) \hp f_3(\binvar)$ for appropriate functions of $f_1, f_2, f_3$ are used in \cite{srivastava2015training, wang2017sort, dong2017more, dauphin2017language, liu2021pay, peng2022branchformer, go2022gswin, pan2022amam, wang2022medical, chrysos2023regularization, chen2024multilinear, fan2024rmt}. For instance, in the Gated Linear Unit (GLU)~\cite{xu2018large}, they assume $f_1 (\binvar) = \bm{0}, f_2 (\binvar) = \bm{W}\binvar$ and $f_3 (\binvar) = \sigma(\bm{V}\binvar)$ for learnt weights $\bm{W}, \bm{V}$. 
In Highway networks~\cite{srivastava2015training} the functions $f_1 (\binvar) = \binvar, f_2 (\binvar) = \binvar - H(\binvar)$ and $f_3 (\binvar) = \sigma (\bm{W}\binvar)$ are used. 

Beyond second-order interactions, polynomial expansions of arbitrary degree have been introduced, which propose various ways to reduce the parameters of \cref{eq:prodpoly_poly_general_eq}. 
A straightforward solution is to assume a low-rank structure in each of the tensors. The CP decomposition along with \cref{lemma:polygan_lemma_hadamard_kr2} is used to obtain the output $\boutvar$ as an $N\myth$ degree expansion~\cite{chrysos2022polynomial}:
\begin{equation}
\begin{split}
    \boutvar = \bm{\beta} + \bm{C}_{1,[1]}^T \binvar + \Big(\bm{C}_{1,[2]}^T \binvar\Big) \hp \Big(\bm{C}_{2,[2]}^T \binvar\Big) + \ldots +  \\ \underbrace{\Big(\bm{C}_{1,[N]}^T \binvar\Big) \hp \ldots \hp \Big(\bm{C}_{N,[N]}^T \binvar\Big)}_{\text{N Hadamard products}}\;,
\end{split}
\label{eq:nosharing_model_no_sharing}
\end{equation}
where $\{ \bm{C}_{i, [j]} \in \realnum^{d\times o} \}_{i, j=1}^N$ are learnable parameters. The index $i$ in the matrices $\bm{C}_{i, [j]}$ corresponds to the $i\myth$ term in the $j\myth$ order interactions. This expression reduces the learnable parameters significantly from exponential to ${\Theta}(N^2 \cdot d \cdot o)$. 

We can reduce further the parameters by assuming that certain factors are shared across different layers. The sharing can be achieved by jointly factorizing all the tensors $\{\bmcal{W}^{[n]} \}_{n=1}^N$. In \pinet~\cite{chrysos2020poly}, various such concrete coupled tensor decompositions are utilized to jointly factorize the tensors. A simple recursive expression of that factorization is exhibited below:
\begin{equation}
    \boutvar_{n} = \left(\bm{A}\matnot{n}^T\binvar\right) \hp \left(\bm{S}\matnot{n}^T \boutvar_{n-1} + \bm{b}\matnot{n}\right)\;,
\label{eq:prodpoly_model2_simplified}\tag{NCP}
\end{equation}
for $n=2,\ldots,N$ with $\boutvar_{1} = \Big(\bm{A}\matnot{1}^T\binvar\Big) \hp \bm{b}\matnot{1}$ and $\boutvar = \bm{C}\boutvar_{N} + \bm{\beta}$. The parameters $\bm{C} \in  \realnum^{o\times k}, \bm{A}\matnot{n} \in  \realnum^{d\times k}, \bm{S}\matnot{n} \in  \realnum^{k\times k}$, $\bm{b}\matnot{n} \in  \realnum^{k}$ for $n=1,\ldots,N$, are learnable with $k \in \naturalnum$ the rank of the decomposition. That is, the final output $\boutvar$ is an $N\myth$ degree polynomial expansion of the input $\binvar$ and is obtained as an affine transformation of the last recursive term $\boutvar_N$. The recursive terms $\{ \boutvar_n \}_{n=1}^N$ provide the intermediate representations. \cref{eq:prodpoly_model2_simplified} includes few core operations that exist in all major deep learning frameworks and as such, it can be easily implemented. \rebuttal{An indicative implementation of \cref{eq:prodpoly_model2_simplified} is added in \cref{ssec:Had_prod_implementations_pinet}}. Note that the learnable parameters are of the order $\Theta (N\cdot d \cdot o)$ owing to the parameters sharing. The recursive formula is obtained by assuming a hierarchical CP decomposition with shared factors as illustrated in \cite{chrysos2020poly}, while different recursive formulations can be obtained if we change the assumptions or the factorization of the terms.

In practice, there are two relaxations over the formulations above: Firstly, often activation functions are used in between the different terms for stabilizing the training~\cite{chrysos2020poly}. Secondly, the aforementioned polynomial expansions are often used as polynomial layers, which are then composed sequentially to obtain the final network. For instance, the second polynomial layer of \cref{eq:prodpoly_model2_simplified} accepts the output of the first polynomial layer of \cref{eq:prodpoly_model2_simplified} as input. That is, the final network is a composition of the polynomial layers. The total degree of expansion is then the product of the degrees of each polynomial layer. The benefit of such a composition is that it can increase the total degree of expansion without increasing the number of layers significantly. For instance, a product of $N$ polynomial layers of degree $2$ results in a $2^N$ total degree. 
\gnote{TODO: possibly illustrate this composition of polynomials with a figure if there is space in the end.}

Let us now demonstrate two additional applications of the aforementioned polynomial expansion of \cref{eq:prodpoly_model2_simplified}.
Fathony et al.~\cite{fathony2021multiplicative} apply an elementwise function $g$ on the output of the term $\left(\bm{A}\matnot{n}^T\binvar\right)$ of \cref{eq:prodpoly_model2_simplified}. They consider two special forms of functions $g$, i.e., the sinusoid and the Gabor filter. In both cases, they illustrate that the final output is a linear combination of Fourier or Gabor bases respectively. Follow-up works further extend the types of filters and add constraints to those~\cite{lindell2022bacon, shekarforoush2022residual}. The idea is to use such networks for approximating low-dimensional functions, e.g., given an $(x, y)$ coordinate of an input image output the RGB values of that coordinate. This implicit representation task has been frequently met across a range of domains, such as computer vision, the last few years.  

The second application concerns the influential work of StyleGAN~\cite{karras2018style}. In StyleGAN, a Hadamard-based generator is introduced for the first time in the Generative Adverarial Nets (GANs)~\cite{goodfellow2014generative}. The Hadamard product is performed in the ADAIN layer~\cite{huang2017arbitrary}, while the generator structure resembles  \cref{eq:prodpoly_model2_simplified}. The authors illustrate how this structure leads to an unsupervised separation of high-level attributes. Follow-up works~\cite{nie2020semi, chrysos2021unsupervised} have further focused on the disentanglement of the high-level attributes using the inductive bias of the Hadamard product.\looseness-1

\gnote{One paragraph about polynomial neural fields here? possibly in the second round.}

Besides the vector-based analysis above, a tailored analysis could be performed depending on the type of data/task. The seminal squeeze-and-Excitation block (\sne)~\cite{hu2018squeeze} relies on an image structure. The output captures pairwise channel correlations of the input $\minvar \in \realnum^{hw \times c}$, where $h$ denotes the height, $w$ the width and $c$ the channels of the image. Notation-wise, the output $\moutvar$ is expressed as:
\begin{equation}
    \moutvar = (\minvar\bm{C}_{\matnot{1}}) \hp r(p(\minvar\bm{C}_{\matnot{1}})\bm{C}_{\matnot{2}})\;.
    \label{eq:nosharing_senet_block} 
\end{equation}
The functions $p, r$ represent the global pooling and the replication of the spatial dimensions respectively. The matrices $\bm{C}_{\matnot{1}} \in \realnum^{c\times \omega}, \bm{C}_{\matnot{2}} \in \realnum^{\omega\times c}$ are learnable. The output of the global pooling $p(\minvar\bm{C}_{\matnot{1}})$ is a vector of size $\omega$. Follow-up works have extended the pairwise channel correlations~\cite{hu2018gather, woo2018cbam, qin2021fcanet}. 
Indicatively, the selective kernel nets~\cite{li2019selective} perform a minor re-parametrization of \cref{eq:nosharing_senet_block} to enable different receptive fields on the features. 
Beyond recognition, \sne{} has been applied to image compression~\cite{gao2021neural}, image super-resolution~\cite{zhang2018image} and denoising~\cite{anwar2019real}, semantic segmentation~\cite{zhang2018context}, medical image segmentation~\cite{ni2022space, wang2022sbdf, ni2022surginet} and point cloud completion~\cite{pan2021variational}.  
Follow-up works have expanded on the role of channel and spatial correlations, e.g., in the context of semantic image synthesis~\cite{tang2020edge} or medical image segmentation~\cite{cheng2020fully, ni2020pyramid}.

High-order interactions can be beneficial beyond the representations of data that we focused above. For instance, high-order interactions are used for expressing non-linear positional encodings~\cite{xiong2016dynamic, chen2021learning}. Positional encodings express the absolute or relative position of an element in a sequence, when the ordering is important. In \cite{wu2022point}, they argue that capturing the relative positioning is more complex in 3D point clouds, since points are unevenly distributed in the space.\looseness-1

Overall, networks capturing higher-order interactions have exhibited state-of-the-art results in 3D modeling~\cite{ploumpis20223d, chrysos2021deep}, pose estimation~\cite{chu2017multi}, image generation~\cite{karras2018style, chrysos2020poly}, audio generation~\cite{wu2022adversarial}, image recognition~\cite{wang2017sort, hu2018squeeze}, segmentation~\cite{yang2022focal} and multimodal representation learning~\cite{wu2019connective, ging2020coot, gao2021fastvideoretrieval, pan2022multimodal}. We hypothesize that this can be partly attributed to the increased expressivity of such networks, as detailed in \cref{sec:elementwise_pr_theoretical_properties}.

\section{Multimodal fusion}   
\label{sec:elementwise_pr_feature_fusion}

\rebuttal{In nature, phenomena can be described using data from various sensors or detectors. For example, humans observe objects and scenes while associating sounds or odors with them. The data distribution from a different sensor is called \emph{modality}~\cite{lahat2015multimodal}. This section focuses on multimodal fusion, which involves tasks that use data from multiple sensors, such as visual and auditory data, and their interactions. The concept of data fusion dates back to at least the seminal work of Hotelling\cite{harold1936relations}. The joint analysis of multiple datasets has been explored since~\cite{cattell1944parallel, vinograde1950canonical}, with significant advancements in the 1970s through factor analysis~\cite{kettenring1971canonical, carroll1970AnalysisOI}. Since then, these ideas have expanded beyond psychometrics and chemometrics.}

\rebuttal{In machine learning, the earliest reference to a multimodal system is the combination of speech and video (image) processing by \cite{petajan1984automatic}. Early studies showed that the interaction between auditory and visual information is non-linear rather than purely additive~\cite{sumby1954visual, mcgurk1976hearing}. Motivated by this complexity, Yuhas et al.~\cite{yuhas1989integration} employed neural networks for multimodal tasks, suggesting that similar frameworks could be beneficial in other multimodal contexts. In recent years, the rapid development of electronic devices and social networks has enabled the unprecedented collection of multimedia sources, often as multimodal data. The Hadamard product has proven particularly useful in multimodal fusion, as demonstrated in visual question answering (VQA) challenges~\cite{teney2018tips}. Suppose we have two input variables, $\sbinvar{I}$ and $\sbinvar{II} \in \realnum^d$, from different modalities. We aim to fuse the information captured across these modalities with many important works, e.g., as mentioned in \cref{tab:hadamard_product_older_important_works_multimodal_fusion}, leveraging Hadamard product for the task.}

\begin{table}[t]
    \centering
    \caption{\rebuttal{\emph{Important} works in the category of multimodal fusion.}}
    {\color{\colReb}
    \begin{tabular}{c c c}
             Task                              & Method                                            & Publication\\\hline
             Visual Question Answering (VQA)   & \cite{antol2015vqa}                               & ICCV'15\\
             VQA                               & MRN~\cite{kim2016multimodal}                      & NeurIPS'16\\
             VQA                               & MUTAN~\cite{ben2017mutan}                         & ICCV'17\\
             Semantic image generation         & SPADE~\cite{park2019semantic}                     & CVPR'19\\
             Knowledge graph completion        & RotatE~\cite{sun2019rotate}                       & ICLR'19\\
             Text-guided image retrieval       & VAL~\cite{chen2020image}                          & CVPR'20\\
             Video-Text Retrieval              & HGR~\cite{chen2020fine}                           & CVPR'20\\
             Vision-and-Language Navigation    & \cite{hong2020language}                           & NeurIPS'20\\
             Conditional generation            & CoPE~\cite{chrysos2021conditional}                & NeurIPS'21\\ \hline
    \end{tabular}
    }
    \label{tab:hadamard_product_older_important_works_multimodal_fusion}
\end{table}

The straightforward idea is to fuse the data from two streams using a function of the form $f_1 (\sbinvar{I}) \hp f_2 (\sbinvar{II})$ for appropriate functions $f_1, f_2$.  In \cite{li2019controllable}, the image and word embeddings are merged in the GAN discriminator using this fusion. In the visual question answering (VQA) of \cite{antol2015vqa, ben2017mutan, anderson2018bottom, lu2019vilbert, lu202012, zhou2020unified, cao2022mobivqa, nguyen2022coarse, xiang2022path} or in video question answering~\cite{gao2021env}, a similar formula is used to fuse the visual and the text embeddings. The same formula with a single Hadamard product emerges in tasks where more than one variables are available~\cite{zhang2017ppr, ramirez2019spectral, georgopoulos2020multilinear, li2021saliency, ramirez2021ladmm, kumaraswamy2021detecting, lu2022onavi}. Lastly, in the hashtag prediction task, the single Hadamard product is preferred over alternative fusion techniques~\cite{durand2020learning}.

\gnote{Note to self: search the follow-up to \cite{kim2016multimodal} or MUTAN~\cite{ben2017mutan}}

The aforementioned works perform feature fusion using a single Hadamard product per modality. However, utilizing ideas from \cref{sec:elementwise_pr_polynomial_nets}, we can capture high-order cross-interactions across modalities. That is, we can capture both the auto- and the cross-correlations of the input variables $\sbinvar{I}, \sbinvar{II}$. A recursive formulation that can capture such higher-order interactions is the following~\cite{chrysos2021conditional}:
%
\begin{equation}
    \boutvar_{n} = \left(\bm{A}\matnot{n, I}^T \sbinvar{I} +  \bm{A}\matnot{n, II}^T \sbinvar{II}\right) * \left(\bm{S}\matnot{n}^T \boutvar_{n-1} + \bm{b}\matnot{n}\right)\;,
    \label{eq:mvp_model2_rec}
\end{equation}
for $n=2,\ldots,N$ with $\boutvar_{1} = \left(\bm{A}\matnot{1, I}^T \sbinvar{I} +  \bm{A}\matnot{1, II}^T \sbinvar{II}\right) * \bm{b}\matnot{1}$. The parameters $\bm{C} \in  \realnum^{o\times k}, \bm{A}\matnot{n, \phi} \in  \realnum^{d\times k}, \bm{S}\matnot{n} \in  \realnum^{k\times k}, \bm{b}\matnot{n} \in \realnum^{k}$ for $\phi=\{I, II\}$ are learnable. Then, the output is $\boutvar = \bm{C}\boutvar_{N} + \bm{\beta}$. That is, the output is an $N\myth$ degree polynomial expansion with respect to the input variables $\sbinvar{I}, \sbinvar{II}$ and is expressed as the affine transformation of the last recursive term $\boutvar_{N}$. This formulation can be thought of as an extension of \cref{eq:prodpoly_model2_simplified} for the case of two input variables. Having separate matrices $\bm{A}\matnot{n, I}, \bm{A}\matnot{n, II}$ associated with each respective input variable is beneficial when the input variables differ substantially in the information they represent, e.g., low-resolution image versus a data-class label. Special forms\footnote{Even though activation functions are often used among the terms of \cref{eq:mvp_model2_rec} in various works, those do not affect the Hadamard product that is the main focus here. Therefore, we will omit such element-wise activation functions to avoid cluttering the notation.} of \cref{eq:mvp_model2_rec} have been considered in the literature~\cite{kim2016multimodal, de2017modulating, nam2017dual, chen2018self, zhou2018end, yu2018beyond, deng2018visual, duke2018generalized, gao2019multi, lee2020maskgan, jiang2020defense, kim2020hypergraph, xia2021visual, akula2021robust, wang2021interpretable, zhan2022bi, bar2023multidiffusion, kim2023dense, zou2024language}.  

In particular, the formulation is often applied in the case of text-to-image generation~\cite{qiao2019mirrorgan, liao2022text}, where one variable captures the text information and another one the input noise, which is sampled from the Gaussian distribution.
Similarly, in the text-guided image manipulation of \cite{li2020manigan}, the formulation of \cref{eq:mvp_model2_rec} is applied with $\{\bm{A}\matnot{n, II} = \bm{0} \}_{n=2}^N$, where $\sbinvar{II}$ denotes the text information and $\sbinvar{I}$ captures the visual information. That is, the text embeddings are explicitly utilized only on the first Hadamard product, while the visual embeddings are used in subsequent Hadamard products.  
In \cite{wu2022text, wu2022language}, the opposite idea was illustrated. That is, $\{\bm{A}\matnot{n, I} = \bm{0} \}_{n=2}^N$, where $\sbinvar{I}$ captures the visual information. 
In \cite{sun2019image}, they synthesize an image given the bounding boxes of the objects (layout) as conditional information. If we denote the input noise, which is sampled from the Gaussian distribution, as $\sbinvar{I}$ and the conditional information (which is the layout) as $\sbinvar{II}$, \cite{sun2019image} considers  $\{\bm{A}\matnot{n, I} = \bm{0} \}_{n=2}^N$. A similar idea is used in \cite{li2016visual} for VQA, where $\sbinvar{I}$ captures the visual information, $\sbinvar{II}$ the text information and $\{\bm{A}\matnot{n, I} = \bm{0} \}_{n=2}^N$.

The seminal work of SPADE~\cite{park2019semantic} can be cast as a form of \cref{eq:mvp_model2_rec} with respect to a single variable similarly to \cite{li2016visual}. This variable corresponds to the semantic layout and is the conditional input in SPADE. If $\sbinvar{II}$ is the conditional input, SPADE assumes that $\bm{A}\matnot{n, I} = \bm{0}$ for $n=2,\ldots,N$ in \cref{eq:mvp_model2_rec}.
Object-shape information~\cite{lv2022semantic} have been added to SPADE to extract additional features from the conditional variable (i.e., the semantic map). An interesting extension is to include additional variables, e.g., to express style information~\cite{zhu2020sean}.  

The recursive form of \cref{eq:mvp_model2_rec} can be extended to facilitate an arbitrary number of variables~\cite{chrysos2021conditional}. For instance, for a third input variable $\sbinvar{III}$, we apply an affine transformation with learnable parameters $\{ \bm{A}\matnot{n, III} \}_{n=1}^N$ and then augment the first term of \cref{eq:mvp_model2_rec} with $\bm{A}\matnot{n, III}^T \sbinvar{III}$. Such a formulation arises in text-guided image generation~\cite{zhang2021cross}, where they capture correlations between the conditional information, i.e., text, the intermediate representations, and the input noise. In the VQA framework of \cite{do2019compact}, they capture the correlations between the image, question and answer representations using a formula of the form $(\bm{A}\matnot{I} \sbinvar{I}) \hp (\bm{A}\matnot{II} \sbinvar{II}) \hp (\bm{A}\matnot{III} \sbinvar{III})$ with $\bm{A}\matnot{I}, \bm{A}\matnot{II}, \bm{A}\matnot{III}$ learnable. This ternary relationship is further extended in \cite{xu2023tfun} for capturing ternary image-text retrieval (TITR). To be precise, they evaluate TITR in recipe retrieval (where the ingredient text, the instruction text and the food image are the three variables) and in fashion search (where the original image, the modified image and the text are the three variables). The same expression is captured in \cite{ruwa2019triple} between text embeddings, visual embeddings and sentimental attributes for sentimental VQA. Note that the aforementioned formula is a special case of \cref{eq:mvp_model2_rec}. 

\gnote{Visual grounding (VG) should be (?) treated as a three variable case, since the objects are extracted from the input image, but then they are processed independently possibly, and we care about the coordinates. For instance in \cite{deng2018visual} they capture a special case with only multiplicative interaction between the image-query fused variable and the object one.}

Alternative formulations have been also considered in the literature. 
For instance, a property of the Hadamard product connecting it with the Fast Fourier Transform (FFT) is exhibited in \cite{fukui2016multimodal} for VQA. That is, the convolution of two vectors $\sbinvar{I}, \sbinvar{II}$ (i.e., word and visual embeddings) can be computed as $\text{FFT}^{-1} (\text{FFT}(\sbinvar{I}) \hp \text{FFT}(\sbinvar{II}))$. Then, they use the Hadamard product twice to fuse the two modalities. In \cite{xiong2016dynamic}, multiple Hadamard products are considered for fusing the visual and the text information for VQA. The co-attention of \cite{ma2018visual} includes two pathways: one that captures high-order correlations of the visual information $\sbinvar{I}$ and one that captures the respective correlations for text $\sbinvar{II}$. Concretely, in the first pathway they use  $\{\bm{A}\matnot{(1), n, II} = \bm{0} \}_{n=2}^N$, while in the second $\{\bm{A}\matnot{(2), n, I} = \bm{0} \}_{n=2}^N$, where we added the $(1), (2)$ indices to separate the pathways. 

Another line of research involves modifying the popular attention mechanism~\cite{vaswani2017attention} for multimodal tasks. For instance, \cite{jiang2022hadamard} modifies the self-attention block using Hadamard product, and applies this for image captioning. A similar modification is used in the attention mechanism of \cite{yang2021auto} for image captioning and VQA. Both matrix multiplications between the (transformed) inputs of the attention are converted to Hadamard products in \cite{liang2022pmacnet}. Attention has also been modified to include a multimodal fusion of visual and text embeddings with Hadamard product in the query and key attributes in \cite{gao2019dynamic}. 


Frequently, visual/text grounding tasks require reasoning about the relationship between objects. For instance, in visual grounding (VG) that idea is to predict the location of the most relevant object given a query expressed in natural language. Message-passing with Hadamard product~\cite{chen2020fine, liu2021region} can be used for such relational reasoning. In \cite{hu2019language}, they construct a message passing graph network, where each object is represented by a node. They use the Hadamard product both for updating the current node features (which they mention as the context features) and merging the visual and the text features. The framework is augmented in \cite{jing2022maintaining} to improve the consistency between different text queries. 
In \cite{gao2020multi}, they rely on message passing with Hadamard product to capture the visual and text interactions. A similar idea with message passing has been used for agent navigation~\cite{hong2020language}. The agent needs to follow instructions, while it has access to the visual scene. The network of the agent uses the Hadamard product to deliver relevant information from the text instructions to the scene. In \cite{rodriguez2021dori}, message passing with the Hadamard product is utilized to capture interactions between the visual and the text embeddings for localizing actions in a video stream.

\gnote{This message passing is equivalent to capturing higher-order interactions I feel.}

The graph structure is also frequently assumed in knowledge graphs beyond the aforementioned message passing. Knowledge graphs consist of a number of triplets $(\bm{h}, \bm{r}, \bm{t})$ that relate a head entity $\bm{h}$ and a tail entity $\bm{t}$ with a relationship $\bm{r}$. Such a triplet can be the following: EPFL (head entity) is located in Switzerland (tail entity). Often, there are missing information in a knowledge graph, e.g., not all the relationships are provided, and we need to perform knowledge graph completion. The Hadamard product holds a key role for the knowledge graph completion of RotatE~\cite{sun2019rotate}. The entities and relationships are mapped to complex space vectors, with $\bm{r}$ also being a unitary vector (i.e., $\left| {r}_i\right| = 1$ with $\bm{r} = [r_1, r_2, \ldots, r_d]$). Then, the relationship in RotatE is modeled as a rotation from the head to the tail entity as $\bm{t} = \bm{h} \hp \bm{r}$. In \cite{le2023knowledge}, they augment RotatE by including convolutional correlations with an additional Hadamard product. The idea of RotatE is extended to include semantic hierarchies in HAKE~\cite{zhang2020learning}. RotatE has also been modified to capture correlations among different triplets~\cite{bai2021prrl}. The work of \cite{zhang2019quaternion} uses the Hadamard product to implement a similar idea with quaternions. SimplE~\cite{kazemi2018simple} requires the Hadamard product to capture correlations between entities from other triplets, while CrossE~\cite{zhang2019interaction} learns "interacting" embeddings that jointly capture the entity-relationship correlations.

Beyond the aforementioned applications, the Hadamard product has been used in medical imaging and neuroimaging, where different modalities are often required for a diagnosis~\cite{croitor2011fusing}. For instance, the fusion of magnetic resonance imaging (MRI) and positron emission tomography (PET) modalities~\cite{shi2017multimodal}, or fusion of MRI and ultrasound images~\cite{el2020ultrasound} have utilized the Hadamard product. We expect that the adoption of deep learning approaches in neuroimaging can lead to further applications of the Hadamard product~\cite{lou2021multimodal}. 
The Hadamard product is used for fusing visual and text embeddings in the medical visual question answering (Med-VQA)~\cite{hasan2018overview} of \cite{zhan2020medical, li2022bi}. 

Another application of the Hadamard product is in remote sensing, where information about the earth are collected from a distance, e.g., airborne or satellite images. 
Even though monitoring the surface of the earth is important, deploying and maintaining sensors is expensive and trade-offs (e.g., images of low spatial resolution) are often made. 
Multimodal fusion can then be used to extract accurate information about complex scenes, e.g., by merging information from different wavelengths/sensors~\cite{dalla2015challenges}. 
In \cite{zhou2020class}, a single Hadamard product is used to guide the image representations using class-information, while multiple Hadamard products are used in \cite{jin2022dasfnet}.  The attention mechanism is modified to include Hadamard product in \cite{liu2020afnet, feng2022icif}, in a similar way as mentioned above for image captioning. 
In \cite{zheng2021gather}, they argue that RGB information can provide useful features, but it also results ambiguity because of the complex texture involved, thus auxiliary information are required. They propose a block in the form of \cref{eq:mvp_model2_rec} (for $N=2$) for the fusion of those modalities.

\gnote{Note to self: TODO: general fusion scheme: \cite{dai2021attentional} } 
\gnote{TODO: Possibly create one table where we have papers for the same applications, e.g. VQA, image captioning, etc together. Or even better a schematic that somehow shows which modalities they require. One major cluster is the text-image one. }

\section{Adaptive modulation}   
\label{sec:elementwise_pr_gating}

Int this section, we focus on the role of the Hadamard product in masking. Frequently, only part of the input should be considered, e.g., in the case of image inpainting we want the occluded part to be filled in using the rest of the input image. Similarly, in image generation we might want to synthesize the same object with different background, or equivalently synthesize the same (human) voice with different auditory backgrounds. We examine below cases where masking is utilized, either in the form of a binary mask or in a soft mask or even as a dynamic modulation of the weights. Concretely, in \cref{ssec:elementwise_pr_masking}, we review various masking and adaptive modulation techniques, while on \cref{ssec:elementwise_pr_rnn} we attend to the recurrent models that perform a dynamic modulation of the inputs. \rebuttal{Few important works on the topic are mentioned in \cref{tab:hadamard_product_older_important_works_adaptive_modulation}.}\looseness-1

\gnote{An alternative writing to the existing: mention that sometimes the mask is external info directly, ..., or computed using second variable, e.g., in \cite{zhang2021text}, or it is computed from the single input. Mention also that the mask might be obtained through an optimization procedure, such as in \cite{borsoi2019super}. }

\begin{table}[t]
    \centering
    \caption{\rebuttal{\emph{Important} works in the category of adaptive modulation.}}
    {\color{\colReb}
    \begin{tabular}{c c c}
             Task                              & Method                                            & Publication\\\hline
             Sequence learning                 & LSTM~\cite{hochreiter1997long}                    & Neural computation'97\\
             Image recognition                 & DropConnect~\cite{wan2013regularization}          & ICML'13\\
             Vision, speech, NLP               & Dropout~\cite{srivastava2014dropout}              & JMLR'14\\
             Sequence learning                 & GRU~\cite{cho2014properties}                      & SSST'14\\
             Image inpainting                  & \cite{pathak2016context}                          & CVPR'16\\
             Image inpainting                  & \cite{yu2018generative}                           & CVPR'18\\
             Language modeling                 & T-Few~\cite{liu2022few}                           & NeurIPS'22\\\hline
    \end{tabular}
    }
    \label{tab:hadamard_product_older_important_works_adaptive_modulation}
\end{table}

\subsection{Masking with Hadamard product}
\label{ssec:elementwise_pr_masking}

\textbf{Binary masks}: In inpainting, a binary mask is often provided as input along with the image to be filled in~\cite{pathak2016context, yu2018generative}. The binary mask indicates which values should be filled in (marked with values $1$), while the rest of the mask attains the value $0$. Then, a Hadamard product is utilized either in the forward pass~\cite{ren2019structureflow, yi2020contextual, jam2021r, wadhwa2021hyperrealistic} or in the backward pass~\cite{yu2018generative}. If the Hadamard product is used in the forward pass, it expresses effectively a branching function for each pixel. That is, if the pixel should remain, it maintains the same value, otherwise a function is applied to that pixel. 
Beyond image inpainting, binary masks have been used in video inpainting~\cite{kim2019deep}, depth completion~\cite{atapour2019veritatem}, image blending~\cite{hoyer2021three}, 3D mesh generation~\cite{pavllo2020convolutional}, image editing~\cite{abdal2020image2stylegan++}. In \cite{tang2020local}, binary semantic class masks are used for semantic-guided scene generation. 
A binary tree mask is constructed in \cite{hui2020linguistic} to restrict the adjacency matrix of the word graph for text-guided image segmentation. 


Beyond the fixed, binary mask, we can consider updating the binary mask, e.g., by partially reducing the masked area~\cite{liu2018image, oh2019onion}. \rebuttal{Pruning the weights of a neural network is an exemplary usecase of binary masks. Mask values of 0 effectively prune the corresponding weights, while values of 1 retain the original weights~\cite{frankle2019lottery, nova2023gradient, iurada2024finding}.}
The seminal work of Dropout~\cite{hinton2012improving} belongs in this category. Dropout aims at reducing overfitting, by masking at random few elements in the intermediate representations. If $\boutvar_n$ denotes the intermediate representations at the $n^{\text{th}}$ layer of a network (cf. \cref{eq:elementwise_pr_mlp}) and $\bm{\gamma} \sim \mathcal{B}$ denotes sampling from the Bernoulli distribution $\mathcal{B}$, then Dropout is expressed as:\looseness-1 
\begin{equation}
    \bm{x}_{n+1} = \bm{\gamma} \hp \boutvar_n\;,
    \label{eq:elementwise_pr_dropout}
\end{equation}
where the $\bm{x}_{n+1}$ is the input for the next layer, i.e., $\boutvar_{n+1} = \sigma (\bm{S}_n \bm{x}_{n+1})\bm$ for appropriate $\sigma, \bm{S}_n$. Numerous extensions have been introduced to Dropout, such as sampling from a Gaussian distribution~\cite{srivastava2014dropout},  adapting the dropout probability~\cite{ba2013adaptive}, or regularizing dropout for representation consistency~\cite{wu2021r}. A principled understanding of Dropout has sparked the interest of the community~\cite{arora2021dropout}. A more thorough overview of the developments on Dropout can be found in related surveys~\cite{labach2019survey}. An idea similar to \cref{eq:elementwise_pr_dropout} was proposed as a feature-wise transformation layer for regularizing the representations~\cite{guo2019attention} or the channels of the representations~\cite{tseng2020cross}.

\rebuttal{Hadamard product is also used for causal language modeling with linear attention. }
Attention and self-attention mechanisms~\cite{vaswani2017attention} have a dramatic influence across different domains the last few years~\cite{radford2018improving, devlin2019bert, han2022survey, touvron2023llama}. The attention mechanism, which was originally introduced for machine transduction, reduced the computational complexity over the previously used recurrent models for such sequence-to-sequence tasks. Given a sequence of length $n$ and the input matrices $\bm{Q} \in \realnum^{n \times d_k}, \bm{K} \in \realnum^{n \times d_k}$ and $\bm{V} \in \realnum^{n\times d_v}$, the attention mechanism is expressed as follows:
\begin{equation}
    \bm{Y}_{SA}= \sigma_s \left( \frac{\bm{Q} \bm{K}^T}{\sqrt{d_k}} \right) \bm{V}\;,
    \label{eq:elementwise_pr_attention}
\end{equation}
where $\sigma_s$ is frequently the row-rise softmax activation function. The self-attention (SA) used in practice in \cite{vaswani2017attention} relies on a single sequence, i.e., the $\bm{Q}, \bm{K}, \bm{V}$ are affine transformations of the common input $\minvar$, e.g., a sequence of word embeddings. That is, the input $\minvar \in \realnum^{n\times d}$ is transformed into $\bm{Q} = \minvar \bm{W}_q , \bm{K} = \minvar \bm{W}_k$ and $\bm{V} = \minvar \bm{W}_v$ with learnable parameters $\{\bm{W}_{\tau} \in \realnum^{d\times d_k} \}_{\tau \in \{q, k\}}$ and $\bm{W}_v \in \realnum^{d\times d_v}$.

\rebuttal{To address the quadratic complexity of self-attention, linear self-attention was proposed~\citep{katharopoulos2020transformers}, where $\sigma_s$ is replaced by an identity matrix. This approach has been adopted in several recent advancements in language models~\citep{sun2023retentive,beck2024xlstm, zhang2024gated}. During language-model training, the task is to predict the next token based on all previous tokens. That would effectively mean that to output a single sentence consisting of $m$ tokens, we would need to make $m$ forward passes, which is very costly. Instead, a masking mechanism can enable to make the prediction in a single forward pass  ensuring that a query cannot access keys from future tokens. Mathematically, this process can be formulated as follows~\footnote{\rebuttal{The nonlinear feature mapping for $\bm{Q}$ and $\bm{K}$ are omitted for notation simplification.}}:
\begin{equation}
    \bm{Y}_{\text{LIN\_SA}}=  \left\{ \left( \frac{\bm{Q} \bm{K}^T}{\sqrt{d_k}} \right)  \hp \bm{R} \right\} \bm{V}\;,
    \label{eq:lin_attention}
\end{equation}
where $\bm{R}$ is the masking matrix, i.e., a lower triangular matrix with entries 1, as present in \cref{fig:elementwise_pr_motivational_fig}(c). To provide further information on masked language modeling, we include an indicative PyTorch implementation in \cref{ssec:Had_prod_implementations_linear_attention_masked}.}

Binary masks, in which the mask is predicted as the output of a network or an optimization process, have naturally emerged in image/video synthesis tasks. The method of \cite{bae2022furrygan} continuously updates the distribution of masks, since they train a mask generator for explicitly disentangling foreground from background in image synthesis. Similarly, in \cite{huang2022layered}, they feed two consecutive frames (of a video) into a network to estimate the foreground/background mask and then condition the synthesis of the new frame into those masks. Another application is outlier detection, where a learnable binary mask is used to filter examples that do not belong to the training distribution~\cite{jewell2022one}. 



\gnote{The next two "categories" below are the gating mechanisms, since the original work mentioning gating is the LSTM, and this has the format below. }

\textbf{Soft masks}: Beyond the aforementioned applications of binary masks, we could consider masks as probabilities, i.e., the probability a pixel is correct or needs to be refined. Each element in this mask accepts a value in the range $[0, 1]$. This mask is referred to as ``soft mask'' henceforth.  

The image inpainting of \cite{xie2019image} uses soft masks to update the regions that have been completed and which need to be refined. 
In the video captioning of \cite{zhou2018end}, the region proposals are converted to a soft mask and then fused with the visual embeddings for determining which regions are relevant for video captioning. In \cite{cong2022anomaly}, the Hadamard product is used to emphasize the anomalous regions in medical images. In \cite{tang2020edge}, an edge-based mask is created to guide the semantic image generation. In the spiking neural networks of \cite{zhao2022backeisnn}, a self-feedback mechanism is inserted using the Hadamard product. 
Various perspectives on soft masks with the utilization of the Hadamard product have also been considered for tumor detection and classification~\cite{carneiro2015weakly, chen2018focus}, text-guided image manipulation~\cite{wu2022language, hou2022feat}, image inpainting~\cite{yu2019free}, semantic image synthesis~\cite{liu2019learning}, deraining~\cite{rai2022fluid}, cancer diagnosis~\cite{chen2020pathomic}, medical imaging super-resolution~\cite{chen2022dynamic}, recommender systems~\cite{ma2019hierarchical} and click-through rate prediction~\cite{wang2021masknet}. 
In the approaches of \cite{drumetz2016blind, borsoi2019super, li2019bilateral}, the scaling matrix is only constrained to contain positive values, but not explicitly constrained in the $[0, 1]$ interval.\looseness-1

Lastly, we consider the case of adaptive modulation of the input, where typically there are two representations and we consider the weighted average of those element-wise. For inputs $\bm{A}$ and $\bm{B}$, this is expressed as:
\begin{equation}
    \bm{Y} = \bm{R} \hp \bm{A} + (\bm{1} - \bm{R}) \hp \bm{B}\,,
\label{eq:elementwise_pr_modulation_weighted_avg}
\end{equation}
where the $\bm{R}$ is the weighting matrix that accepts values in the $[0, 1]$ interval. 
\cref{eq:elementwise_pr_modulation_weighted_avg} is also frequently used in multimodal fusion (complementary to the aforementioned techniques in \cref{sec:elementwise_pr_feature_fusion}). In the image manipulation of \cite{zhang2021text}, the textual instructions are computing the weighting matrix, and the visual and text embeddings are fused using the \cref{eq:elementwise_pr_modulation_weighted_avg}. 
In \cite{zhu2022cefusion}, this equation is used for multimodal medical image fusion, where there are two source images and the weighting matrix is obtained as the output of the decoder. In the face swapping of \cite{li2019faceshifter}, the attribute embeddings and the identity embeddings are fused using a weighting matrix that is computed from the visual embeddings. \cref{eq:elementwise_pr_modulation_weighted_avg} has been used for fine-grained image generation of the foreground/background~\cite{yang2017lr}. The work of \cite{singh2019finegan} relies on the previous fine-grained generation to perform disentangled image generation with weak supervision. The RelTransformer~\cite{chen2022reltransformer} captures the relationships between objects (entities) using \cref{eq:elementwise_pr_modulation_weighted_avg}. Masking through the Hadamard product is used for reducing the unimodal biases in VQA~\cite{cadene2019rubi}. Concretely, the mask is used to alter the predictions of the network, in order to force the network to avoid learning biases based on the question. 
In both text-guided images synthesis~\cite{zhu2019dm}, and multispectral imaging~\cite{ji2022unified}, \cref{eq:elementwise_pr_modulation_weighted_avg} plays an important role.

Beyond the idea of masking representations, masking the parameters of a network has been utilized across a range of tasks. A major use case considers various subnetworks of the overall architecture, e.g., by using a binary mask to make certain weights zero. For instance, ProbMask~\cite{zhou2021effective} proposes to sparsify the weights using a Hadamard product with a mask. 
The influential work of DropConnect~\cite{wan2013regularization} induces sparsity on the weights for regularizing the network. Inspired by Dropout (i.e., \cref{eq:elementwise_pr_dropout}), in every iteration a binary masked is sampled from the Bernoulli distribution and a different subnetwork is obtained.  Blockout~\cite{murdock2016blockout} modifies DropConnect by inserting a heuristic for determining the binary mask instead of randomly sampling it. 
In \cite{mallya2018piggyback}, the binary masking of the weights enables the network to handle multiple tasks. In the few-shot approach of \cite{liu2022few}, the adaptation of the weights with a learned vector (that is multiplied with the regular weights via Hadamard product) enables effective fine-tuning for new tasks. In the case of federated learning, a binary mask enables each client to obtain a personalized model~\cite{dai2022dispfl}. 
In \cite{yan2018spatial, kong2018graph, tan2022multi}, a Hadamard product between a weighting matrix and the edges of a graph is performed for importance weighting. 
In \cite{fu2021hyperspectral}, they scale the representation coefficients for hyperspectral imaging. In \cite{mackay2019self}, a scaled affine transformation of the hyperparameters operates as the weighting matrix for hyperparameter optimization.  
Lastly, in \cite{cheung2019superposition}, masking enables a single network to have multiple convolutional kernels for multi-task learning. 

\gnote{TODO: subsection of other usages of masking? Pruning and bias reduction below. Possibly this is a subsection in the end of this section with the title: Masking the weights/parameters}

\subsection{Recurrent neural networks with Hadamard product}
\label{ssec:elementwise_pr_rnn}

\rebuttal{The Long Short-Term Memory (LSTM) network is the most widely used recurrent neural network (RNN), with the Gated Recurrent Unit (GRU) as a notable variant. This section highlights their structures and key applications. For more details, see surveys like \cite{yu2019review}.}

The Long Short-Term Memory (LSTM)~\cite{hochreiter1997long} was originally introduced as a remedy to the vanishing/exploding gradients of recurrent models trained with back-propagation. 
The structure of an LSTM block is expressed as follows:

\begin{equation}
    \begin{split}
        \bm{\nu}_t = \sigma_g (\bm{W}_z \binvar_t + \bm{U}_z \bm{h}_{t-1} + \bm{b}_z) \;, \\
        \bm{r}_t = \sigma_g (\bm{W}_r \binvar_t + \bm{U}_r \bm{h}_{t-1} + \bm{b}_r) \;, \\
        \bm{o}_t = \sigma_g (\bm{W}_o \binvar_t + \bm{U}_o \bm{h}_{t-1} + \bm{b}_o) \;, \\
        \bm{c}_t = \sigma_c (\bm{W}_c \binvar_t + \bm{U}_c \bm{h}_{t-1} + \bm{b}_c) \;, \\
        \hat{\bm{h}}_t = \hat{\bm{h}}_{t-1} \hp \bm{\nu}_t + \bm{r}_t \hp \bm{c}_t \;, \\
        \bm{h}_t = \bm{o}_t \hp \phi_h (\hat{\bm{h}}_t)\;,
    \end{split}
    \label{eq:lstm}\tag{LSTM}
\end{equation}
for every time step $t > 0$, where $\bm{h}_0=\bm{0}$, and $\hat{\bm{h}}_0 =\bm{0}$. The activation functions $\sigma_c, \phi_c$ are often selected as hyperbolic tangent function (tanh), while $\sigma_g$ denotes a sigmoid. The parameters $\bm{W}_{\tau}, \bm{U}_{\tau}, \bm{b}_{\tau}$ for $\tau \in \{\nu, r, c, o\}$ are learnable. The output vector $\bm{h}_t$ can then be used for prediction. The symbols $\bm{\nu}_t, \bm{r}_t, \bm{o}_t$ are known as the forget, update and the output gates respectively. The gates enable information to be stored for longer periods of time. For instance, if the forget gate $\bm{\nu}_t$ is close to $\bm{0}$, the hidden state of the the previous time steps (i.e., $\hat{\bm{h}}_{t-1}$) will be "forgotten".

Gated Recurrent Unit (GRU) was originally introduced as a block for machine translation~\cite{cho2014properties, chung2014empirical}. GRU is a lightweight variant of LSTM and it is expressed as:

\begin{equation}
    \begin{split}
        \bm{\nu}_t = \sigma_g (\bm{W}_z \binvar_t + \bm{U}_z \bm{h}_{t-1} + \bm{b}_z) \;, \\
        \bm{r}_t = \sigma_g (\bm{W}_r \binvar_t + \bm{U}_r \bm{h}_{t-1} + \bm{b}_r) \;, \\
        \hat{\bm{h}}_t = \phi_h \left( \bm{W}_h \binvar_t + \bm{U}_h (\bm{r}_t \hp \bm{h}_{t-1}) + \bm{b}_h \right) \;, \\
        \bm{h}_t = \bm{\nu}_t \hp \bm{h}_{t-1} + (1 - \bm{\nu}_t) \hp \hat{\bm{h}}_t\;,
    \end{split}
    \label{eq:gru}\tag{GRU}
\end{equation}
for every time step $t > 0$, where $\bm{h}_0=\bm{0}$ and $\binvar_t$ is the input. The parameters $\bm{W}_{\tau}, \bm{U}_{\tau}, \bm{b}_{\tau}$ for $\tau \in \{\nu, r, h\}$ are learnable. The symbols $\bm{\nu}_t, \bm{r}_t$ are known as the update and the reset gates respectively. Further modifications have been proposed, such as removing the reset gate~\cite{ravanelli2018light}, or modifying the matrix multiplications to convolutions~\cite{wang2020deep}.

 Overall, LSTM, GRU, as well as more recent recurrent models, such as the RHN~\cite{zilly2017recurrent} and IndRNN~\cite{li2018independently}, rely on the Hadamard product to modulate the amount of information that is allowed through for sequence modelling.

\section{Computationally efficient operator} 
\label{sec:elementwise_pr_efficient_op}

Hadamard product has also been used for reducing the communication cost or accelerating well-established operations. For instance, in \cite{hyeon2022fedpara} they use a low-rank re-parametrization of the weights of the network to reduce the communication cost of federated learning. Then, they use Hadamard product to capture richer interactions without increasing the communication cost. 
In \cite{dong2017more}, they propose to augment the convolutional layer by a low-cost layer along with ReLU activation function to induce sparsity, and then use Hadamard product with the regular representation. They show how the new layer can lead to acceleration over the traditional \resnet.  
Beyond the aforementioned cases, we illustrate below how the Hadamard product has been used in self-attention variants to reduce the computational complexity, or how it is used in practice to implement certain activation functions.

\subsection{Self-attention variants} 

\rebuttal{As a reminder, the self-attention (SA) as expressed in \cref{eq:elementwise_pr_attention} is a hugely influential mechanism.}
Despite its empirical success, SA suffers from quadratic complexity with respect to the input size. One approach to reduce the computational cost is to replace the matrix multiplications with Hadamard products~\cite{babiloni2021poly, zhai2021attention, wu2021fastformer, jiang2022swinbts, hua2022transformer, ma2022mega, yang2025gated}. In fact, if we ignore the softmax\footnote{A number of recent works have experimentally replaced softmax. For instance, in Primer~\cite{so2021searching} the activation function search indicates that a quadratic ReLU performs favorably in large-scale experiments.}, we can construct a new module relying on Hadamard products that obtains exactly the same element-wise relationship between the input elements $X_{i, j}$~\cite{babiloni2021poly}. Concretely, given the input $\minvar \in \realnum^{n\times d}$ (same as the SA above), the poly-SA can be expressed as:

\begin{equation}
    \bm{Y}_{\scalebox{.55}{\text{Poly-SA}}} = \left\{\Psi\left((\minvar\bm{W}_{1}) \hp (\minvar\bm{W}_{2}) \right) \hp \minvar\right\} \bm{W}_{3}\;,
    \label{eqn:fast_nl}
\end{equation}
where $\Psi\colon\mathbb{R}^{n \times d} \to \realnum^{n \times d}$ is an average pooling on the second dimension followed by a replication on the same dimension. The matrices $\bm{W}_{1},\bm{W}_{2} \in \realnum^{d \times d}$ and $\bm{W}_{3} \in \realnum^{d \times d_v}$ are learnable\footnote{The weights $\bm{W}_{1},\bm{W}_{2}, \bm{W}_{3}$ are different than $\bm{W}_{Q},\bm{W}_{K}, \bm{W}_{V}$, e.g., the dimensions of the corresponding matrices differ. See the original publication~\cite{babiloni2021poly} for further details.}. The computational cost is reduced from quadratic to linear with respect to the input size. Intuitively, this is achieved due to the form of the Hadamard product, which only requires a single element of each input matrix to compute the corresponding output. On the contrary, the matrix multiplications that \cref{eq:elementwise_pr_attention} relies on, require access to the whole row/column to compute an output element.

\subsection{Multiplicative activation functions}

Beyond the theoretical and the technical benefits of using the Hadamard product, we argue that there are also benefits on the software side. As we demonstrate below the Hadamard product is actively used for implementing certain activation functions. 
Even though ReLU is widely used with convolutional or fully-connected networks, it has two core drawbacks: its derivative at zero and the zero update signal if the neuron is inactive. Various alternative activation functions have been proposed to mitigate those drawbacks.  
Several of those element-wise activation functions~\cite{ramachandran2017searching, misra2019mish, zhu2021logish} contain the form $f_1(\invar)\cdot f_2(\invar)$. For instance, in SiLU/Swish~\cite{hendrycks2016gaussian, ramachandran2017searching} $f_1$ is the identity function, while $f_2$ is the sigmoid function. Interestingly, Swish activation function approaches ReLU for certain values of its hyper-parameter, while Swish was discovered using a large-scale search on activation functions. In Mish~\cite{misra2019mish}, $f_1$ is the identity function, while $f_2(\invar) = \text{tanh} \left(\log (1 + e^{\invar}) \right)$. The implementation of such activation functions relies on the Hadamard product in popular deep learning frameworks, such as PyTorch\footnote{The related lines in the source code of PyTorch for SiLU and Mish are \url{https://bit.ly/3Vh0Ok0} and \url{https://bit.ly/3GQz6qD} respectively.}. 
%
\rebuttal{Variants of those activation functions are used in strong-performing language models, including Llama~\cite{touvron2023llama}.}
Beyond the aforementioned activation functions, polynomial activation functions offer certain theoretical~\cite{livni2014computational, zhang2019encrypted} and empirical~\cite{lokhande2020generating} benefits, while they can be expressed using Hadamard products for an efficient implementation. 
Interestingly, the polynomial activation functions also capture high-order correlations as we detail in \cref{sec:elementwise_pr_polynomial_nets}. (Piecewise) polynomial functions~\cite{ahlberg1967theory} have interesting properties, but are outside of the scope of this work.\looseness-1

\section{Theoretical understanding}
\label{sec:elementwise_pr_theoretical_properties}

The empirical success of the Hadamard product across a broad range of learning applications has fostered the investigation of its theoretical properties in machine learning. In this section, we recap the theoretical analyses of this operator from the aspects of its expressivity, spectral bias, generalization, robustness, extrapolation, and verification, when used in neural networks.
\rebuttal{These theoretical properties have immediate impact on practical applications. For instance, the benefit of learning high-frequency functions faster is significant in specific applications, e.g., in the StyleGAN generator. In addition, we believe that the benefits in terms of expressivity and extrapolation properties can be important considerations for future applications.}

\textbf{Expressivity:}
In deep learning theory, the expressivity of neural networks is a foundational question asking which class the network can characterize~\citep{raghu2017expressive}.
\citep{Jayakumar2020Multiplicative} studies the role of multiplicative operations by showing that replacing the linear layer in neural networks with such operations can enlarge hypotheses space when a single Hadamard product is present. 
In the traditional neural network, each neuron is defined as \cref{eq:elementwise_pr_mlp}.
 As a comparison, a quadratic neuron (with a single Hadamard product) has the following formula:
 $
    \outvar
    =\sigma{\left(
 (\binvar^\top \vw_1 + b_1)
(\binvar^\top \vw_2 + b_2)
+
(\binvar \hadp \binvar)^\top \vw_3 + b_3
 \right)}$.  
~\citep{10.1016/j.neunet.2020.01.007} claim that the quadratic networks with ReLU activation function can approximate radial function in a more efficient and compact way compared to traditional neural network.
~\citep{10.1016/j.neunet.2020.01.007} also provably demonstrate the global universal approximation of quadratic network with ReLU activation function. Beyond radial function,~\citet{fan2021expressivity} shows the higher expressivity of quadratic networks compared to traditional networks. Specifically, based on spline theory, \citet{fan2021expressivity} proves that with ReLU activation function, quadratic networks are more expressive than traditional neural networks. Even though when the ReLU activation in traditional neural networks is replaced with quadratic activation, it still has poor expressivity compared to the quadratic networks with ReLU activation.

\textbf{Spectral bias:}
Modern neural networks have demonstrated strong power in fitting complex functions or even random labels~\citep{zhang2017understanding}. Moreover, the excellent generalization performance and the ability to avoid over-fitting seem to be inconsistent with the model complexity in generalization theory. An intriguing phenomenon, called spectral bias, aims to unravel this mystery. Spectral bias supports that neural networks demonstrate a learning bias towards low-complexity functions~\citep{rahaman2019spectral}. Indeed, the spectral bias provides an intuitive explanation for the disagreement between over-fitting and model complexity since neural networks first fit low-complexity functions during training and thus belong to the function class of low complexity.

A line of works has tried to explain such spectral bias theoretically utilizing the neural tangent kernel (NTK). The NTK is widely used to analyze neural networks from a range of aspects, e.g., convergence, generalization, and memorization~\citep{NEURIPS2018_5a4be1fa,cao2019towards,nguyen2021tight}.
Specifically, the NTK is defined as the limit of the inner product of the gradients (with respect to the parameters), when we assume an infinite width $k$ and a specific Gaussian initialization. That is, the NTK matrix is expressed as 
$K(\vz, \vz^{\prime})=\lim _{k \rightarrow \infty}\left\langle\nabla_{\vw} f_{\vw}(\vz), \nabla_{\vw} f_{\vw}\left(\vz^{\prime}\right)\right\rangle
\,,
$
where $\vw$ denotes all vectorized weights of the network.
Using NTK, the spectral bias of a two-layer feed-forward ReLU neural network, which is expressed as
$
\outvar
={\vc}^\top \sigma{(\bm{A}\matnot{1}^\top \binvar)}$, with ReLU activation function $\sigma$, was verified. That is, the two-layer feed-forward network learned low-frequency functions faster. On the contrary, if we add a single Hadamard product, the phenomenon changes, both empirically and theoretically. To be more precise, let us express this network as:
 $
\outvar
=
{\vc}^\top
\left(
\sigma{(\bm{A}\matnot{1}^\top \binvar)}
\hadp 
\sigma{(\bm{A}\matnot{2}^\top\binvar)}
\right)\;.
$
Then, this network admits a slower eigenvalue decay of the respective NTK matrix, resulting in a faster learning towards high-frequency functions~\cite{choraria2022the}. This analysis was recently extended to networks with more Hadamard products in ~\cite{wu2022extrapolation}. 

Even though the NTK has several limitations, e.g., the infinite width or the lazy training assumption, it is one of the most valuable and influential tools in deep learning theory for the understanding of neural networks. In practice, it has been used for providing actionable insights for the training and design of neural networks~\cite{zhu2022generalization}. Lastly, the analysis in \citet{choraria2022the} relies on the assumption that the input data is uniformly distributed on the sphere, a future step would be extending the result to non-uniform data distributions.

\textbf{Generalization/Robustness}
Beyond the expressivity and the inductive bias of the networks with Hadamard product, a key question concerns the generalization performance. Indeed, the generalization error and the robustness to adversarial perturbations of polynomial networks of the form of \cref{eq:prodpoly_model2_simplified}, i.e. a particular parametrization of the networks with Hadamard product, were recently explored~\citet{zhenyu2022controlling}. The Rademacher complexity~\citet{shalev2014understanding}, which is often used to characterize the generalization error by assessing how well the function class can correlate with random labels, is used in the theorem below.

\overlengthPAMI{
The Rademacher complexity, which we define below, is commonly used to characterize the generalization error by assessing how well the function class can correlate with random labels.
\begin{definition} [Empirical Rademacher Complexity, Eq.26.4 of \citet{shalev2014understanding}]
\label{def:rademacher}
Suppose $Z = \left \{ \bm{z}_{1}, \ldots ,\bm{z}_{|\mathcal{Z}|} \right \} \subseteq \realnum^d$
and $\left \{\eta_{j}:j = 1,\ldots, |\mathcal{Z}| \right \}$ be
independent Rademacher random variables with values uniformly in
$\left \{ -1, +1 \right \}$. The Empirical Rademacher complexity of
a class of real-valued
functions $\mathcal{F}$ over $\realnum^d$.
is defined as:
\begin{equation*} 
        \mathcal{R}_{Z}(\mathcal{F}) \coloneqq \mathbb{E}_{\eta} \bigg[
\sup_{f\in\mathcal{F}}\frac{1}{|\mathcal{Z}|}\sum_{j=1}^{|\mathcal{Z}|}\eta_{j}f(\bm{z}_{j})
            \bigg]\,.
    \end{equation*}
\end{definition}
}

The Empirical Rademacher Complexity of the polynomial networks $f(\bm{z})$ in \cref{eq:prodpoly_model2_simplified} with 1-D output (i.e., $\bm{C} \in  \realnum^{1 \times k}$) excluding the bias term has been studied.
\begin{theorem}[Theorem 3 of \citep{zhenyu2022controlling}]
\label{theorem_NCP_RC_Linf} 
    Suppose that
    $\|\bm{z}_j\|_\infty \leq 1$ for all $j=1, \ldots, |\mathcal{Z}|$. Define the matrix $\Phi
    \coloneqq (\bm{A}\matnot{N}\bullet \bm{S}\matnot{N})  \prod_{i=1}^{N-1} \bm{I} \otimes \bm{A}\matnot{i} \bullet \bm{S}\matnot{i}$ where $\bullet$ symbolizes the face-splitting product (which can be thought of as the row-wise Khatri-Rao product).
    Consider the class of functions:    
    $
   \mathcal{F} \coloneqq \left \{ f(\bm{z}): \|\bm{C}\|_\infty \leq \mu, \left \|\Phi
   \right \|_\infty \leq \lambda \right \}\,.
    $
    The Empirical Rademacher Complexity of $\mathcal{F}$ with respect to $Z$ can be bounded as: $
        \mathcal{R}_{Z}(\mathcal{F}) \leq 
        2 \mu \lambda \sqrt{\frac{2N\log(d)}{|\mathcal{Z}|}}
        \,.$
\end{theorem}

Note that \cref{theorem_NCP_RC_Linf} connects the Rademacher complexity bounds to the operator norms of the weight matrix. Thus, one can improve the generalization capacity and prevent overfitting by constraining such operator norms as a regularization term.

Another property we are interested is the Lipschitz constant, which  intuitively measures how much the output of the network changes when there is a (small) perturbation in the inputs. The Lipschitz constant has often been explored to measure the robustness to (bounded) perturbations of the input~\citep{cisse2017parseval,tsuzuku2018lipschitz,virmaux2018lipschitz}. Recently, the Lipschitz constant of the polynomial networks has been investigated. 

\overlengthPAMI{

Another property we are interested is the Lipschitz constant, which 
refers to the smallest $K$ in \cref{def:lip}. Intuitively, the Lipschitz constant measures how much the output of the network changes when there is a (small) perturbation in the input of the network. The Lipschitz constant often measures how robust the network is under some (bounded) perturbations of the input~\citep{cisse2017parseval,tsuzuku2018lipschitz,virmaux2018lipschitz}. 

\begin{definition}[Lipschitz continuous]
\label{def:lip}
    Given two normed spaces ($\mathcal{Z}$, $\| \cdot \|_{\mathcal{Z}}$) and ($\mathcal{Y}$, $\| \cdot \|_{\mathcal{Y}}$), a function $f$ : $\mathcal{Z}\rightarrow \mathcal{Y}$ is called Lipschitz continuous with constant $K \geq 0$ if for all $z_1, z_2$ in $\mathcal{Z}$:
    $
    \| f(z_{1}) - f(z_{2}) \|_{\mathcal{Y}} \leq K \| z_{1} - z_{2} \|_{\mathcal{Z}}\,.
    $
\end{definition}
Recently, the Lipschitz constant of the polynomial networks has been investigated.
}

\begin{theorem}[Theorem 4 of \citet{zhenyu2022controlling}]
\label{theorem_CCP_LC_Linf} 
Consider the polynomial networks $f(\bm{z})$ in \cref{eq:prodpoly_model2_simplified} with 1-D output (i.e., $\bm{C} \in  \realnum^{o}$) excluding the bias term. The Lipschitz constant (with respect to the $\ell_\infty$-norm) of $f(\bm{z})$ restricted to the set $\{\bm{z} \in \realnum^d: \|\bm{z} \|_\infty \leq 1 \}$ can be bounded as:
    $
        \text{Lip}_\infty(f) \leq N \| \bm{C} \|_\infty \prod_{n=1}^{N}(\|\bm{A}\matnot{n} \|_\infty \|\bm{S}\matnot{n}\|_\infty)\,.
    $
\end{theorem}
Notably, one can see the relationship between the Lipschitz constant bounds and the operator norms of the weight. Thus, constraining such norms allows for improving the robustness of the network. One open question is whether similar bounds can be obtained for broader classes of functions with Hadamard product, or whether tighter results on the Lipschitz constant and the generalization error can further lead to new regularization schemes. 


\textbf{Verification:}
Neural networks are sensitive to malicious perturbations even in the input data space. Frequently, those attacks are constructed by adding small perturbations to the original input. The perturbations can negatively affect machine learning systems such as image recognition~\cite{szegedy2013intriguing, goodfellow2015explaining}. From a security point of view, it becomes increasingly important to verify the robustness of neural networks against such performance degradation. Verification of neural network refers to investigate the relationship between the inputs and outputs. For example, one can verify that whether adding small perturbation to the input can produce a change of the output which leads to misclassification.

Verification has mostly focused on ReLU networks owing to their widespread use. Recently, the first verification method for the special category of polynomial networks (e.g., the parametrization of \cref{eq:prodpoly_model2_simplified}) was introduced~\cite{rocamora2022sound}. Particularly, they focus on the adversarial robustness restricted to neighborhoods defined with $L_\infty$ norm, which is mathematically formulated as the following problem: Suppose the network function $\bm{f}
$ classifies the input $\bm{z} \in {[0,1]}^{d}$ into a class $c$, such that $c = \argmax{\bm{f}(\bm{z})}$.
Given an observation $\bm{z}_{0}$, denote 
the correct class as $t = \argmax{\bm{f}(\bm{z}_{0})}$ and define a set that contains the neighbourhood of $\bm{z}_{0}$ as follows:
$C_{\text{in}}  = \{\bm{z}: ||\bm{z}-\bm{z}_{0}||_{\infty} \leq \epsilon, z_{i} \in [0,1], \forall i \in [d]\}$.
Verification aims to certify whether every input in 
the set $C_{\text{in}}$ is classified as $t$.
This can be further reformulated as a constrained optimization problem. For every adversarial class $\gamma \neq t = \argmax{\bm{f}(\bm{z}_{0})}$, we need to solve
\begin{equation}
        \begin{aligned}
        \min_{\bm{z}} \quad & g(\bm{z}) = f(\bm{z})_t - f(\bm{z})_{\gamma} \quad
        \textrm{s.t.} \quad %
        \bm{z} \in \mathcal{C_{\text{in}}}\,.\\
        \end{aligned}
\label{eq:verif_problem}
    \end{equation}
If the optimal value $v^{*} = g(\bm{z}^{*})$ 
satisfies $v^{*} > 0$, then the robustness is verified for the adversarial class ${\gamma}$.

Given the non-convexity of the problem, common first order algorithms, such as gradient descent, are not suitable, as there is no guarantee of the equivalence of the stationary point and the global minimum. The branch and bound (BaB)~\citep{Land_Doig1960bab,wang2021beta} has been recently used for solving \cref{eq:verif_problem} for ReLU networks. BaB divides the original problem into sub-problems by splitting the input domain depending on the different branches of ReLU activation function.
In each subproblem, the upper and lower bounds of the minimum are computed. If the lower bound of a subproblem is greater than the current global upper bound, then it can be discarded. Such recursive iteration finishes when the difference between the global upper and lower bound is less than a small constant $\epsilon$.

The existing methods of verification highly rely on the piecewise linearity property of ReLU so that it can not be trivially extended to neural network with Hadamard product. To this end, ~\citet{rocamora2022sound} proposes a variant of BaB algorithm called VPN. Specifically, VPN relies on the twice-differentiable nature of \cref{eq:prodpoly_model2_simplified}, and utilizes the $\alpha$-convexification to compute the lower bound of the minima of each subset. Empirical results demonstrate that this method can obtain much tighter bounds than baselines. 


\textbf{Extrapolation:}
Neural networks exhibit a stellar performance when the test data is sampled from the same underlying distribution as the training data is, i.e., frequently referred to as in-distribution performance. However, neural networks have been observed to have difficulty in learning simple arithmetic problems or fitting simple analytically-known functions outside of the in-distribution regime~\citep{saxton2019analysing,sahoo2018learning}. Indeed, we need to understand when and how a neural network extrapolates outside of the in-distribution regime. Recent work shows that two-layer feedforward neural networks with ReLU activation function extrapolate to linear function~\citep{xu2021how}. Furthermore, only knowing the information of the degree of the extrapolation function is not enough. Naturally, one might ask under which condition neural network can achieve successful extrapolation. To this end,~\citet{xu2021how} provably provides the condition for neural networks to extrapolate the linear target function exactly. 
 However, feature and label usually admit non-linear relationship in real-world.
Intuitively, the polynomial networks discussed in \cref{eq:prodpoly_model2_simplified} have the capacity to extrapolate to non-linear function, as shown in \cref{thm:extra}.
 \begin{theorem}[Theorem 4 of \citet{wu2022extrapolation}]
 \label{thm:extra}
Suppose we train $N$-degree ($N\ge2$) polynomial network $f
$
on $\lbrace (\vz_i, y_i)\rbrace_{i=1}^{|\mathcal{Z}|}$ with the squared loss in the NTK regime. For any direction $\vv \in \R^d$ that satisfies $\| \bm v \|_2 =\max\lbrace\| \vz_{i} \|^2\rbrace$, let 
$\vz = (t + h) \vv$ with $t>1$ and $h>0$ be the extrapolation data points, the output $f( (t+ h) \vv)$  follows a $\gamma$-degree ($\gamma \leq N$) function with respect to $h$.
\end{theorem}
Note that the above theorem holds for polynomial nets with the commonly-used ReLU activation function, or without activation function. One future step is to study the extrapolation of neural networks with polynomial activation function and compare their extrapolation ability with polynomial networks. In addition, investigating more complicated out-of-distribution problems in the real-world, e.g., domain adaptation might be a promising topic.

\section{Open problems and future directions}   
\label{sec:elementwise_pr_discussion}

\rebuttal{The widespread use of Hadamard product highlights its key role in deep learning. Our taxonomy on diverse applications enables cross-pollination of ideas or properties. Below, we summarize key points, future research areas, and current limitations.}

\cref{sec:elementwise_pr_polynomial_nets} exhibits how the Hadamard product is an essential component in capturing high-order interactions in practice. One question that has yet to be answered is under which circumstances it is beneficial to augment the neural network to capture high-order interactions theoretically. Another question is whether such high-order correlations can be used as standalone, i.e., whether the polynomial expansions can reach the performance of state-of-the-art neural networks without using activation functions. We hypothesize that the performance of the polynomial expansions can be further improved by using strong regularization schemes. Additionally, the strong performance of polynomial networks when used in conjunction with activation functions provides an interesting direction of research on the role of activation functions on its own right. 

\cref{sec:elementwise_pr_feature_fusion} develops how the Hadamard product has been widely used for feature fusion. However, to our knowledge there is no theoretical or empirical evidence of how it fares with respect to other feature fusion methods used in the literature, e.g., concatenation, cross-attention or tensor decompositions. Given that feature fusion is an important aspect and an increasing number of multimodal applications are emerging, we believe that both the theoretical properties and the empirical comparison of feature fusion schemes will be beneficial. 

The role of the Hadamard product in adaptive modulation has emerged in various forms in the literature as exhibited in \cref{sec:elementwise_pr_gating}. An interesting question is whether we could further enforce the adaptivity through feedback on the previous modulation. Another interesting avenue is whether this adaptivity is essential on trillion-parameter models or whether the sheer volume of parameter combinations suffices for capturing this representation modulation. 

\cref{sec:elementwise_pr_efficient_op} develops certain cases where the Hadamard product has been used to replace matrix multiplication. One interesting question is whether the self-attention variants with Hadamard product can still perform on par with self-attention on tasks with imbalanced or limited data. In addition, we believe that applications in federated learning, where efficient communication are required, might result in new applications of the Hadamard product. 

Even though some of the theoretical properties of the Hadamard product have been explored the last few years, there are many properties that have received little attention to date. For instance, the impact of Hadamard product into the loss landscape remains largely underexplored.  
So far, the properties of local minima and saddle points of training the first layer of a network with a Hadamard product  (e.g., \cref{eq:prodpoly_model2_simplified} with $N=2$ and general convex loss functions) have been investigated~\citep{du2018power}.
Extending the result to networks with all layers trained or considering neural networks with multiple Hadamard products as well as non-linear activation functions are significant future directions. Similarly, a tighter Lipschitz constant estimation could have benefits in the robustness of such networks. In addition, we expect that further studies on the non-linear extrapolation properties of the networks with Hadamard product will further illustrate their differences with feed-forward neural networks. 
Another promising direction is the study of the rank of Hadamard product both algebraically, and empirically. Understanding this rank is expected to provide a strong guidance on which applications can benefit the most from Hadamard products~\cite{kressner2017recompression}. In addition, often the Hadamard product is used in conjunction with tailored normalization schemes, e.g., in \stylegan, or in SPADE. Even though there is no theoretical grounding for this yet, we hypothesize that this might be beneficial for regularizing the high-order interactions. 

A key topic that has been underexplored in machine learning is how the Hadamard product fares with respect to the noise. Especially in the case of multimodal fusion, it is likely that the data from different modalities have different types and levels of noise. The exact positioning of the Hadamard product on the architecture is an interesting topic, especially in the context of multimodal fusion. It is possible to capture interactions both closer to the data-side, in the intermediate representations between the modalities or closer to the output representation.

\bibliographystyle{IEEEtran}
\bibliography{refs}

\vskip -2\baselineskip plus -1fil
    \begin{IEEEbiography}[{\includegraphics[width=1in,height=1.25in,clip,keepaspectratio]{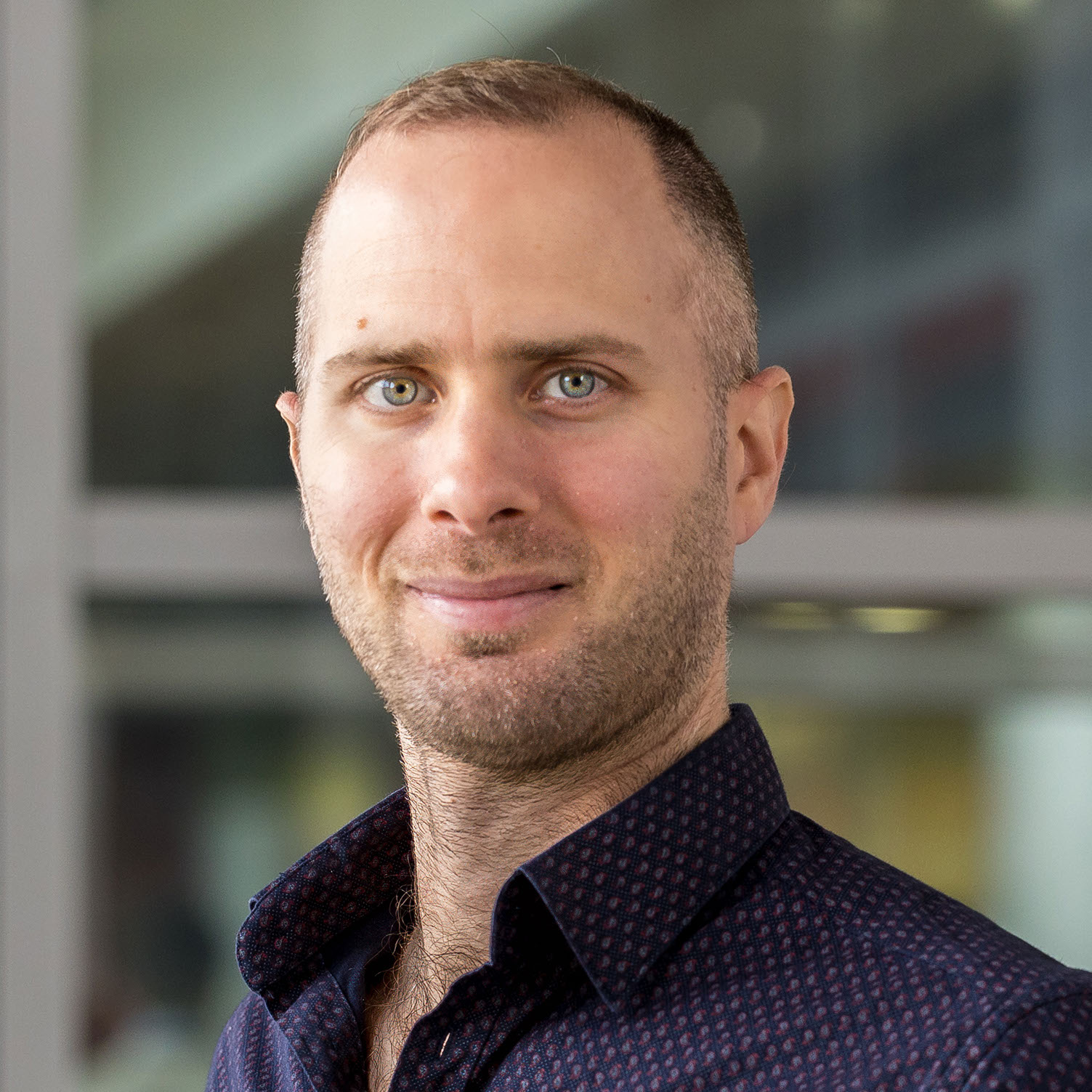}}]{Grigorios G. Chrysos} is an Assistant Professor at the University of Wisconsin-Madison. Before that, Grigorios was a postdoctoral fellow at EPFL following the completion of his PhD at Imperial College London. Previously, he graduated from the National Technical University of Athens. His research interests lie in multilinear algebra, architecture design, generative models, and designing models robust to noise and out-of-distribution samples. Grigorios publishes regularly on ML conferences (NeurIPS, ICML, ICLR), or  top-tier journals. Grigorios serves as an Associate Editor for TMLR and an Area Chair for ML conferences (NeurIPS, ICLR, ICML). \looseness-1
\end{IEEEbiography}

\vskip -2\baselineskip plus -1fil
\begin{IEEEbiography}[{\includegraphics[width=1in,height=1.25in,clip,keepaspectratio]{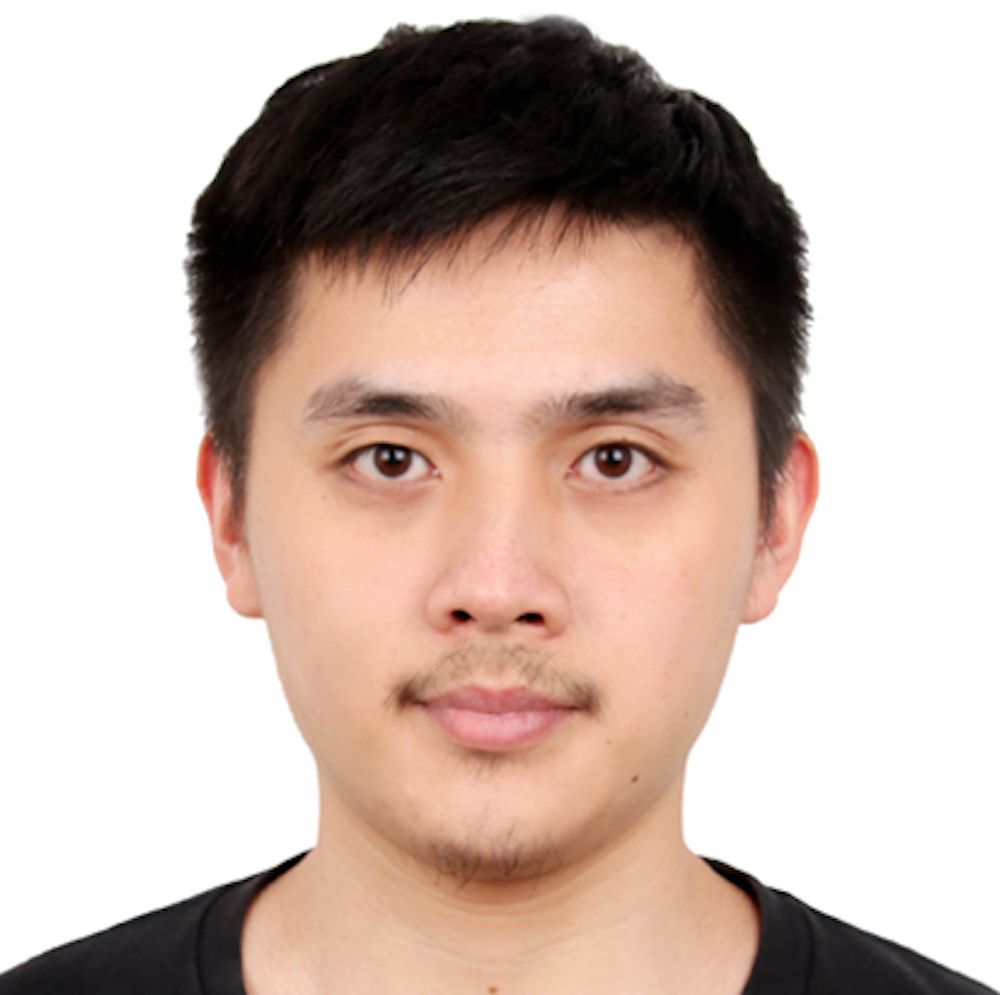}}]{Yongtao Wu} is a PhD student in the Laboratory for Information and Inference Systems (Lions) at École Polytechnique Fédérale de Lausanne (EPFL). Previously, he obtained his BEng diploma in Telecommunication from Sun Yat-sen University and MSc diploma in Machine Learning from KTH Royal Institute of Technology. His research interests are in machine learning with a focus on deep learning theory, polynomial networks, and optimization.
\end{IEEEbiography}

\vskip -2\baselineskip plus -1fil
\begin{IEEEbiography}[{\includegraphics[width=1in,height=1.25in,clip,keepaspectratio]{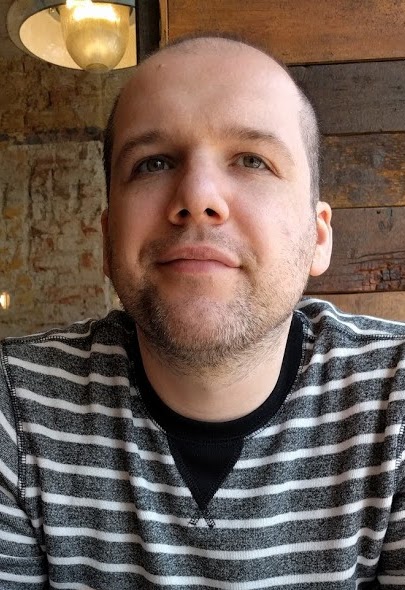}}]{Razvan Pascanu} is a research scientist at DeepMind, UK. He finished his Master Students in 2009 at Jacobs University Bremen, and obtained his PhD in 2014 from University of Montreal, Canada. His research interests focus on Deep Learning and Deep Reinforcement Learning, publishing several impactful works at ML conferences. He organized workshops on the topic of Continual Learning (Neurips'18, ICML'20), Graph Neural Networks (ICLR'19) as well as tutorial on polynomial neural networks (CVPR'22). Additionally he organized the EEML summer school yearly from 2018 onwards, and was program chair for Conference on Lifelong Learning Agents (2022) and Learning on Graphs (2022). 
\end{IEEEbiography}

\vskip -2\baselineskip plus -1fil
\begin{IEEEbiography}[{\includegraphics[width=1in,height=1.25in,clip,keepaspectratio]{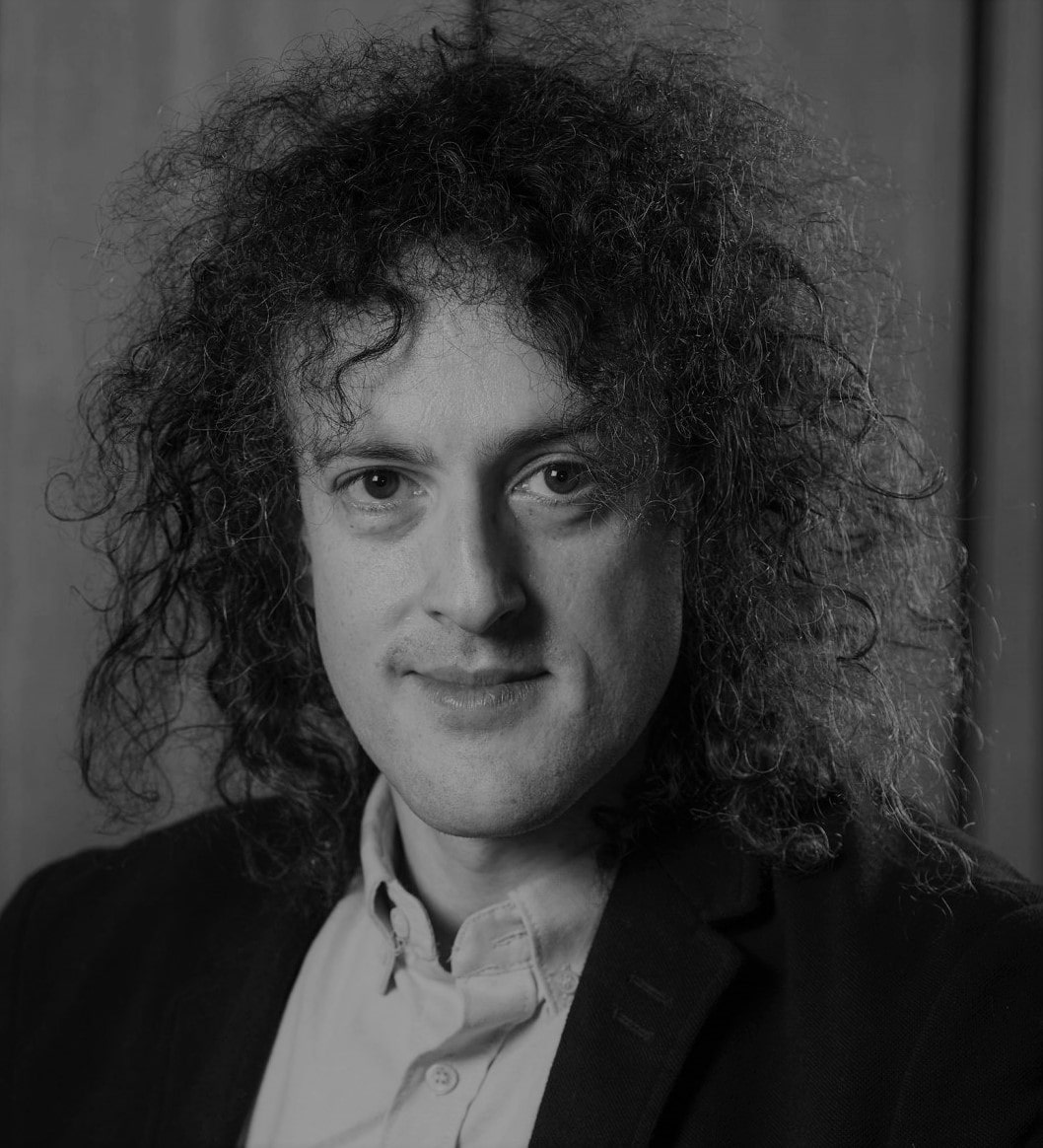}}]{Philip H.S. Torr}
received the PhD degree from Oxford University, U.K. After working for another three years at Oxford as a research fellow, he worked for six years in Microsoft Research, first in Redmond, then in Cambridge, founding the vision side of the Machine Learning and Perception Group. He then became a professor in Computer Vision and Machine Learning at Oxford Brookes University, U.K. He is currently a professor at Oxford University.
\end{IEEEbiography}

\vskip -2\baselineskip plus -1fil
\begin{IEEEbiography}[{\includegraphics[width=1in,height=1.25in,clip,keepaspectratio]{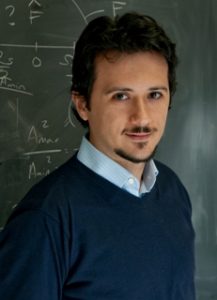}}]{Volkan Cevher} is an Associate Professor at EPFL. He received the B.Sc. (valedictorian) in EE from Bilkent University in Ankara, Turkey, in 1999 and the Ph.D. in ECE from the Georgia Institute of Technology in Atlanta, GA in 2005. 
His research interests include machine learning, signal processing theory, optimization theory and methods, and information theory. Dr. Cevher is an ELLIS fellow and was the recipient of the Google Faculty Research award in 2018, the IEEE Signal Processing Society Best Paper Award in 2016, a Best Paper Award at CAMSAP in 2015, a Best Paper Award at SPARS in 2009, and an ERC CG in 2016 as well as an ERC StG in 2011. 
\end{IEEEbiography}

\clearpage

\renewcommand{\thefigure}{S\arabic{figure}} 
\renewcommand{\thetable}{S\arabic{table}} 
\renewcommand{\thetheorem}{S\arabic{theorem}} 
\renewcommand{\theequation}{S\arabic{equation}}
\renewcommand{\thesection}{S.\arabic{section}}
\renewcommand{\thesubsection}{S.\arabic{section}.\arabic{subsection}}
\renewcommand{\appendixname}{sec.}
\setcounter{section}{0}
\setcounter{equation}{0}

\section{Preliminaries}
\label{sec:elementwise_pr_preliminaries}

\begin{figure*}[!tbh]
    \centering
    \includegraphics[width=0.7\textwidth]{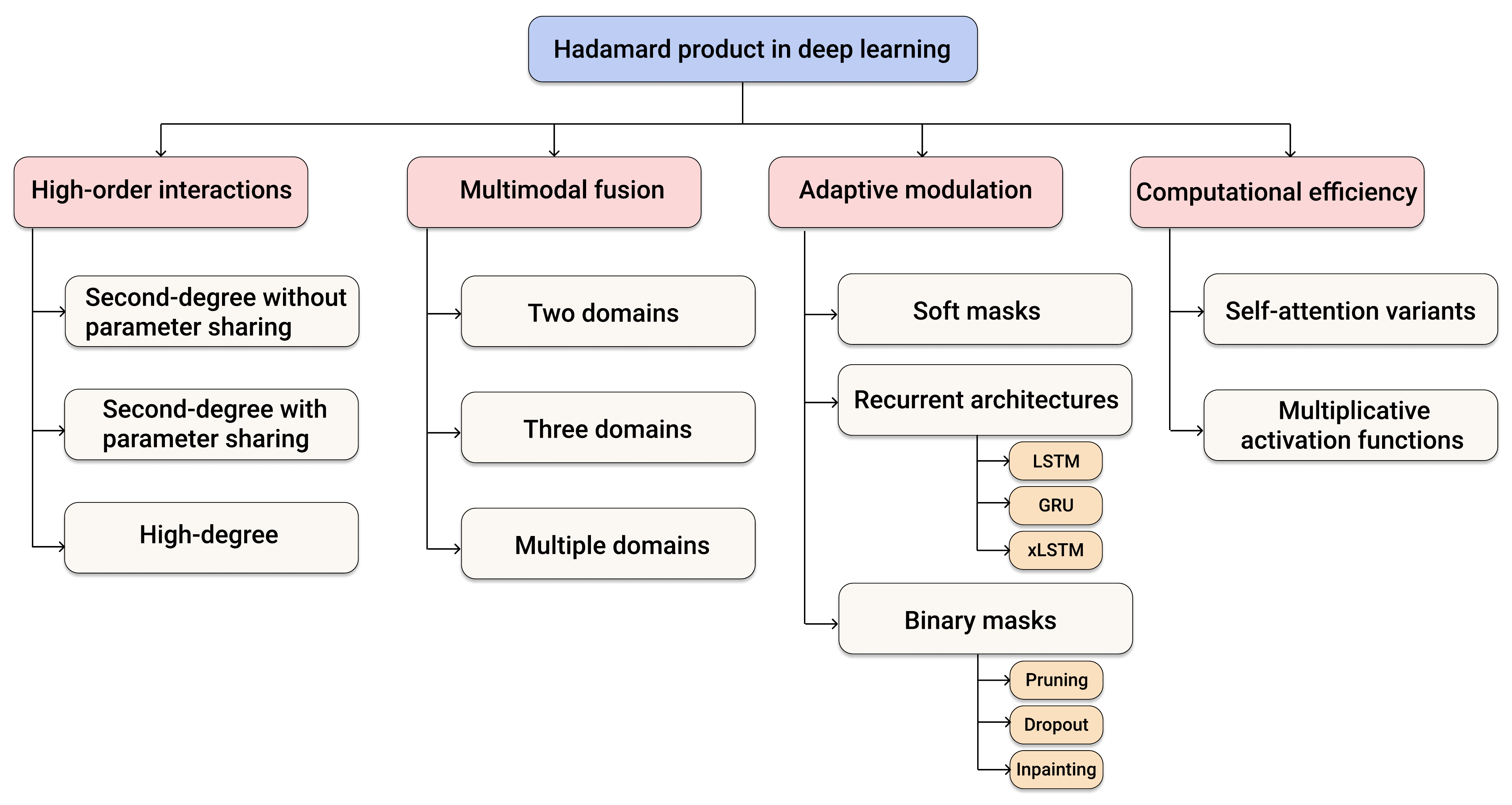}
    \caption{\rebuttal{Taxonomy of Hadamard product in deep learning. The category of high-order interactions is often divided by the degree of interactions, with a more fine-grained taxonomy being whether there is parameter-sharing, i.e., \cref{eq:nosharing_model_no_sharing} vs \cref{eq:prodpoly_model2_simplified}. Similarly, in multimodal fusion, the number of domains is a fundamental separation, with techniques such as VQA belonging to two domains. We believe in the following years, works will increasingly focus on multiple domains for general applications, which simulates how humans perceive and process multiple domains.}}
    \label{fig:elementwise_pr_figure_taxonomy_visual}
\end{figure*}

Below, we provide a detailed overview of the notation in \cref{ssec:elementwise_pr_notation}, and then we introduce few mathematical properties of the Hadamard product in \cref{ssec:elementwise_pr_mathematical_properties}. Lastly, we briefly review the form of feed-forward deep networks in \cref{ssec:elementwise_pr_deep_learning}. For readers that prefer to understand the high-level ideas of this work, this section can be skipped at a first reading.

\subsection{Notation}
\label{ssec:elementwise_pr_notation}
Scalars are denoted with plain letters, e.g., M, i. Matrices (vectors) are denoted by uppercase (lowercase) boldface letters e.g., $\bm{X}$, ($\bm{x}$). The symbol $\bm{0}$ denotes a vector or matrix, where each element is zero.
We follow the standard Bachmann-Landau asymptotic notation, e.g., $\mathcal{O}, {\Theta}$. 

\textbf{Tensors} are core components in this paper. Tensors represent data structures of multiple dimensions and are denoted by boldface, calligraphic letters, e.g.,  $\bmcal{X}$. The
number of indices needed to reference the elements of a tensor is the \textit{order} of the tensor. For instance, a tensor $\bmcal{X} \in
 \realnum^{I_1 \times I_2 \times I_3}$ has order $3$. Each element of an $M\myth$-order tensor $\bmcal{X} \in  \realnum^{I_1 \times I_2 \ldots \times I_M}$ is referenced by $M$ indices, i.e., $(\bmcal{X})_{i_{1}, i_{2}, \ldots, i_{M}} \doteq x_{i_{1}, i_{2}, \ldots, i_{M}}$.  

Let us now provide three core definitions: the mode-m vector product, the CP decomposition and the Hadamard product. The first two definitions are required for transforming various high-order interactions into practical networks, e.g., in polynomial expansions. The definitions below are operations on tensors $\bmcal{X}, \bmcal{Y} \in
 \realnum^{I_1 \times I_2 \times \cdots \times I_M}$ with $I_{m} \in \naturalnum$ for $m=1,2,\ldots,M$.

\begin{definition}[mode-$m$ vector product]
    The \emph{mode-$m$ vector product} of $\bmcal{X}$ with a vector $\bm{u} \in \realnum^{I_m}$, denoted by $\bmcal{X} \times_{m} \bm{u} \in \realnum^{I_{1}\times I_{2}\times\cdots\times I_{m-1}  \times I_{m+1} \times \cdots \times I_{M}} $, results in a tensor of order $M-1$, which is defined element-wise as:
    \begin{equation}
        (\bmcal{X} \times_{m} \bm{u})_{i_1, \ldots, i_{m-1}, i_{m+1},
        \ldots, i_{M}} = \sum_{i_m=1}^{I_m} x_{i_1, i_2, \ldots, i_{M}} u_{i_m}\;.  \nonumber
    \end{equation}
    Furthermore,  we denote $\bmcal{X} \times_{1} \bm{u}^{(1)} \times_{2} \bm{u}^{(2)} \times_{3}  \cdots \times_{M} \bm{u}^{(M)}  \doteq  \bmcal{X} \prod_{m=1}^m \times_{m} \bm{u}^{(m)}$.
\end{definition}

\begin{definition}[CP decomposition]
    The Canonical-Polyadic (CP) decomposition~\cite{hitchcock1927tensor_polyadic, carroll1970AnalysisOI} of $\bmcal{X}$  aims to find the vectors 
    $ \bm{u}^{(1)}_{r} , \bm{u}^{(2)}_{r}, \ldots, \bm{u}^{(M)}_{r}$ for $r\in \{1, 2, \ldots, R\}$,  such that:
    \begin{equation}
        \bmcal{X} = \sum_{r=1}^{R}  \underbrace{\bm{u}^{(1)}_{r} \circ \bm{u}^{(2)}_{r} \circ \ldots \circ \bm{u}^{(M)}_{r} }_{\text{rank-1 components (tensors)}}\;, \nonumber
    \end{equation}
    where $\circ$ symbolizes the outer product. The vectors $\{ \bm{u}^{(m)}_{r} \}_{m=1}^M$ can be collected into the matrices $\left\{ \bm{U}\matnot{m} \doteq \left[ \bm{u}^{(m)}_{1}, \bm{u}^{(m)}_{2}, \ldots, \bm{u}^{(m)}_{R} \right]\right\}_{m=1}^M$.  In this work, we are mostly interested in the matricized format of the CP decomposition which is expressed as follows (for mode-1 unfolding):
    $\bm{X}_{(1)} \doteq \bm{U}\matnot{1} \left( \bigodot_{m=M}^2 \bm{U}\matnot{m}\right)^T$. The symbol $\bigodot$ denotes the Khatri-Rao product~\cite{khatri1968solutions} of a set of matrices.
\end{definition}

\subsection{Mathematical properties of the Hadamard product}
\label{ssec:elementwise_pr_mathematical_properties}

We mention below a few fundamental properties of the Hadamard product. Whenever possible, we express the property for general tensor structures, which means the property also holds for the special cases of matrices and vectors. 
Firstly, the identity element in the case of Hadamard product is the tensor $\bmcal{I}$, which is a tensor with every element taking the value $1$.
\begin{lemma}
    The Hadamard product is commutative, associative and distributive over addition. That is for the tensors $\bmcal{X}, \bmcal{Y}, \bmcal{Z} \in
 \realnum^{I_1 \times I_2 \times \cdots \times I_M}$ it holds that:
    \begin{equation}
    \begin{split}
        \bmcal{X} \hadp \bmcal{Y} =  \bmcal{Y} \hadp \bmcal{X}, \\
       \bmcal{X} \hadp \left( \bmcal{Y} \hadp \bmcal{Z} \right) = \left(\bmcal{X} \hadp \bmcal{Y} \right) \hadp \bmcal{Z},\\
       \bmcal{X} \hadp \left( \bmcal{Y} + \bmcal{Z} \right) = \bmcal{X} \hadp \bmcal{Y} +  \bmcal{X} \hadp \bmcal{Z}.
    \end{split}
    \end{equation}
\end{lemma}

A plethora of interesting properties of the Hadamard product are applicable between matrices~\cite{styan1973hadamard, ando1995majorization} or tensors~\cite{sun2018some, xu2019some}. 
Below, we review few such properties that are relevant to machine learning (ML), focusing on the cases of matrices. Firstly, the Schur product Theorem~\cite{schur1911bemerkungen} states that the Hadamard product of two positive definite matrices is also a positive definite matrix. Follow-up works have focused on understanding the rank of the Hadamard product~\cite{Ballantine1968-lz, horn2020rank}, its minimum eigenvalue~\cite{li2007lower} or links with the regular matrix multiplication~\cite{caro2012connection}. 

The connection of the Hadamard product and the Khatri-Rao product enables the efficient implementation of polynomial networks in deep learning frameworks, e.g. PyTorch. That is, the Khatri-Rao product is connected to well-known tensor decompositions, such as the CP decomposition and as the Lemma below shows, we can replace the Khatri-Rao product, e.g., in the CP decomposition above, with the Hadamard product. 

\begin{lemma}[Lemma 2 in \cite{chrysos2019polygan}] 
    Let us denote $\{\bm{A}_{\nu} \in \realnum^{I_{\nu} \times K} \}_{\nu=1}^N$, $\{\bm{B}_{\nu} \in \realnum^{I_{\nu} \times L} \}_{\nu=1}^N$ two sets of $N \geq 2$ matrices, then it holds that
    \begin{equation}
        (\bigodot_{\nu=1}^N \bm{A}_{\nu})^T \cdot (\bigodot_{\nu=1}^N \bm{B}_{\nu}) = (\bm{A}_1^T \cdot \bm{B}_1) * (\bm{A}_2^T \cdot \bm{B}_2) * \ldots * (\bm{A}_N^T \cdot \bm{B}_N). \nonumber
    \end{equation}
    \label{lemma:polygan_lemma_hadamard_kr2}
\end{lemma}




\subsection{Deep Learning paradigm}
\label{ssec:elementwise_pr_deep_learning}
The deep learning paradigm refers to the broad concept of learning deep neural networks to express complex functions. The expressivity of deep neural networks has been demonstrated both empirically and theoretically across a range of important applications. To make this survey self-contained, we will review the fundamental expression of a DNN below. 

In the feed-forward neural networks, the input is used in the first (few) layer(s) and then a simple recursive formulation is followed for processing the previous layers' outputs. The simple form of this recursive form is provided below for a network with $N$ layers and input $\binvar \in \realnum^d$: 
\begin{equation}
    \boutvar_{n} = \sigma \left(\bm{S}\matnot{n}^T \boutvar_{n-1} + \bm{b}\matnot{n}\right)\;,
    \label{eq:elementwise_pr_mlp}
\end{equation}
for $n=1,\ldots,N$ with $\boutvar_{0} = \binvar$. The output $\boutvar \in \realnum^o$ is then expressed as an affine transformation on the last recursive term, i.e., $\boutvar = \bm{C}\boutvar_{N} + \bm{\beta}$. The parameters $\bm{\beta} \in \realnum^o, \bm{C} \in  \realnum^{o\times k}, \bm{S}\matnot{n} \in  \realnum^{k\times k}$, $\bm{b}\matnot{n} \in  \realnum^{k}$ for $n=1,\ldots,N$ are learnable. The symbol $\sigma$ represents an activation function (e.g., Rectified Linear Unit (ReLU), hyperbolic tangent (tanh), sigmoid), which induces non-linearity to the network. 
In other words, the intermediate representations $\boutvar_n$ can express non-linear functions of the inputs.  

In practice, extensions of \cref{eq:elementwise_pr_mlp} to facilitate components such as normalization schemes~\cite{ioffe2015batch} or residual connections~\cite{he2016deep}, can be included with a minor re-parametrization. For instance, the influential residual block can be expressed as $\boutvar_{n} = \sigma \left(\bm{S}\matnot{n}^T \boutvar_{n-1} + \bm{b}\matnot{n}\right) + \boutvar_{n-1}$, where essentially only an identity of the output from the previous layer is added. 
In the rest of the manuscript, unless mentioned explicitly otherwise $\sigma$ denotes an element-wise activation function, e.g., sigmoid or ReLU, below. Similarly, in the linear layers $\bm{W}\binvar + \bm{\beta}$ with learnable parameters $\bm{W}, \bm{\beta}$ and input $\binvar$ the bias term $\bm{\beta}$ might be omitted for simplicity.

\newpage
\onecolumn
\section{Practical implementation}
\label{sec:Had_prod_implementations_with_hadamard_product}


\rebuttal{A list of public implementations of works cited in the survey can be found below in \cref{tab:hadamard_product_indicative_author_implementations}. In addition, below we reference few core implementations using the popular PyTorch. These are mostly for instructional purposes, while we recommend to the interested reader to follow the public implementation of the authors to reproduce the results of the respective paper.}

\begin{table*}[!t]
    \centering
    \caption{\rebuttal{Indicative list of public implementations (frequently from authors) on many of the works cited in the survey.}}
    { \color{\colReb}
        \begin{tabular}{c@{\hspace{0.4cm}}cl}
        \toprule
        \emph{Citation} & \emph{Authors' implementation} & \emph{Implementation} \\
        \midrule[\heavyrulewidth]
        
        \cite{hu2018squeeze} & \checkmark & \url{https://github.com/hujie-frank/SENet} \\ \hline
        \cite{park2019semantic} & \checkmark & \url{https://github.com/NVlabs/SPADE} \\ \hline
        \cite{pathak2016context} & \checkmark & \url{https://github.com/pathak22/context-encoder} \\ \hline
        \cite{yu2018generative} & \checkmark & \url{https://github.com/JiahuiYu/generative_inpainting} \\ \hline
        \cite{dauphin2017language} &  & \url{https://github.com/anantzoid/Language-Modeling-GatedCNN} \\ \hline
        \cite{hua2022transformer} &  & \url{https://github.com/lucidrains/FLASH-pytorch} \\ \hline
        \cite{chen2024multilinear} & \checkmark & \url{https://github.com/Allencheng97/Multilinear_Operator_Networks} \\ \hline
        \cite{fan2024rmt} & \checkmark & \url{https://github.com/qhfan/RMT} \\ \hline
        \cite{bar2023multidiffusion} & \checkmark & \url{https://github.com/omerbt/MultiDiffusion} \\ \hline
        \cite{guerreiro2024layoutflow} & \checkmark & \url{https://github.com/JulianGuerreiro/LayoutFlow} \\ \hline
        \cite{qin2024hgrn} & \checkmark & \url{https://github.com/OpenNLPLab/HGRN2} \\ \hline
        \cite{ma2024rewrite} & \checkmark & \url{https://github.com/ma-xu/Rewrite-the-Stars} \\ \hline
        \cite{chrysos2021conditional} & \checkmark & \url{https://github.com/grigorisg9gr/polynomial_nets_for_conditional_generation} \\ \hline
        \cite{wu2022extrapolation} & \checkmark & \url{https://github.com/LIONS-EPFL/pntk} \\ \hline
        \cite{liu2021pay} &  & \url{https://github.com/lucidrains/g-mlp-pytorch} \\ \hline
        \cite{chrysos2023regularization} & \checkmark & \url{https://github.com/grigorisg9gr/regularized_polynomials} \\ \hline
        \cite{chrysos2022polynomial} & \checkmark & \url{https://github.com/grigorisg9gr/polynomials-for-augmenting-NNs} \\ \hline
        \cite{fathony2021multiplicative} & \checkmark & \url{https://github.com/boschresearch/multiplicative-filter-networks} \\ \hline
        \cite{woo2018cbam} & \checkmark & \url{https://github.com/Jongchan/attention-module} \\ \hline
        \cite{qin2021fcanet} & \checkmark & \url{https://github.com/cfzd/FcaNet} \\ \hline
        \cite{li2019selective} & \checkmark & \url{https://github.com/implus/SKNet} \\ \hline
        \cite{zhang2018image} & \checkmark & \url{https://github.com/yulunzhang/RCAN} \\ \hline
        \cite{anwar2019real} & \checkmark & \url{https://github.com/saeed-anwar/RIDNet} \\ \hline
        \cite{pan2021variational} & \checkmark & \url{https://github.com/paul007pl/VRCNet} \\ \hline
        \cite{tang2020edge} & \checkmark & \url{https://github.com/Ha0Tang/ECGAN} \\ \hline
        \cite{chen2021learning} & \checkmark & \url{https://github.com/woodfrog/vse_infty} \\ \hline
        \cite{chrysos2021deep} & \checkmark & \url{https://github.com/grigorisg9gr/polynomial_nets} \\ \hline
        \cite{yang2022focal} & \checkmark & \url{https://github.com/microsoft/FocalNet} \\ \hline
        \cite{wu2019connective} & \checkmark & \url{https://github.com/AmingWu/CCN} \\ \hline
        \cite{ging2020coot} & \checkmark & \url{https://github.com/simon-ging/coot-videotext} \\ \hline
        \cite{kim2016multimodal} & \checkmark & \url{https://github.com/jnhwkim/nips-mrn-vqa} \\ \hline
        \cite{ben2017mutan} & \checkmark & \url{https://github.com/Cadene/vqa.pytorch} \\ \hline
        \cite{chen2020image} & \checkmark & \url{https://github.com/yanbeic/VAL} \\ \hline
        \cite{hong2020language} & \checkmark & \url{https://github.com/YicongHong/Entity-Graph-VLN} \\ \hline
        \cite{do2019compact} & \checkmark & \url{https://github.com/aioz-ai/ICCV19_VQA-CTI} \\ \hline
        \cite{rodriguez2021dori} & \checkmark & \url{https://github.com/crodriguezo/DORi} \\ \hline
        \cite{zhang2019quaternion} & \checkmark & \url{https://github.com/cheungdaven/QuatE} \\ \hline
        \cite{kazemi2018simple} & \checkmark & \url{https://github.com/Mehran-k/SimplE} \\ \hline
        \cite{zhang2019interaction} & \checkmark & \url{https://github.com/wencolani/CrossE} \\ \hline
        \cite{zhan2020medical} & \checkmark & \url{https://github.com/Awenbocc/med-vqa} \\ \hline
        \cite{liu2020afnet} & \checkmark & \url{https://github.com/athauna/AFNet} \\ \hline
        \cite{srivastava2014dropout} & \checkmark & \url{https://github.com/nitishsrivastava/deepnet} \\ \hline
        \cite{wadhwa2021hyperrealistic} & \checkmark & \url{https://github.com/GouravWadhwa/Hypergraphs-Image-Inpainting} \\ \hline
        \cite{hoyer2021three} & \checkmark & \url{https://github.com/lhoyer/improving_segmentation_with_selfsupervised_depth} \\ \hline
        \cite{pavllo2020convolutional} & \checkmark & \url{https://github.com/dariopavllo/convmesh} \\ \hline
        \cite{hui2020linguistic} & \checkmark & \url{https://github.com/spyflying/LSCM-Refseg} \\ \hline
        \cite{liu2018image} & \checkmark & \url{https://github.com/NVIDIA/partialconv} \\ \hline
        \cite{sun2023retentive} & \checkmark & \url{https://github.com/microsoft/unilm/tree/master/retnet} \\ \hline
        \cite{huang2022layered} & \checkmark & \url{https://github.com/Gabriel-Huang/Layered-Controllable-Video-Generation} \\ \hline
        \cite{xie2019image} &  & \url{https://github.com/Vious/LBAM_Pytorch} \\ \hline
        \cite{hou2022feat} &  & \url{https://github.com/Psarpei/GanVinci} \\ \hline
        \cite{ma2019hierarchical} & \checkmark & \url{https://github.com/allenjack/HGN} \\ \hline
        \cite{li2019faceshifter} &  & \url{https://github.com/justin4ai/hearnet-pytorch} \\ \hline
        \cite{cadene2019rubi} & \checkmark & \url{https://github.com/cdancette/rubi.bootstrap.pytorch} \\ \hline
        \cite{zhou2021effective} & \checkmark & \url{https://github.com/x-zho14/ProbMask-official} \\ \hline
        \cite{mallya2018piggyback} & \checkmark & \url{https://github.com/arunmallya/piggyback} \\ \hline
        \cite{zilly2017recurrent} & \checkmark & \url{https://github.com/jzilly/RecurrentHighwayNetworks} \\ \hline
        \cite{wu2021fastformer} & \checkmark & \url{https://github.com/wuch15/Fastformer} \\ \hline
        \cite{yang2025gated} & \checkmark & \url{https://github.com/NVlabs/GatedDeltaNet} \\ \hline
        \cite{rocamora2022sound} & \checkmark & \url{https://github.com/megaelius/PNVerification} \\ 
        \bottomrule
        \end{tabular}
    }
    \label{tab:hadamard_product_indicative_author_implementations}
\end{table*}

\subsection{PyTorch example of linear attention for language modeling}
\label{ssec:Had_prod_implementations_linear_attention_masked}

\rebuttal{We paste below an instructional implementation of language modeling with linear attention.}

\begin{lstlisting}[language=python]
from torch import nn
import torch

def causal_mask(seq_len):
    """
    Creates a lower-triangular matrix of shape [seq_len, seq_len] with element one for masking
    """
    mask = torch.tril(torch.ones(seq_len, seq_len))
    return mask  # [seq_len, seq_len]

class LinearSelfAttention(nn.Module):
    """
    A simple linear self-attention block with causal masking applied by Hadamard product.
    """
    def __init__(self, hidden_dim):
        super(LinearSelfAttention, self).__init__()
        self.W_q = nn.Linear(hidden_dim, hidden_dim)
        self.W_k = nn.Linear(hidden_dim, hidden_dim)
        self.W_v = nn.Linear(hidden_dim, hidden_dim)

    def forward(self, z):
        """
        Input z: a batch of sequences with embedding [batch_size, seq_len, hidden_dim]
        Returns:
        output: [batch_size, seq_len, hidden_dim]
        """
        batch_size, seq_len, _ = z.size()

        # Compute queries, keys, values
        Q = self.W_q(z)  # [batch_size, seq_len, hidden_dim]
        K = self.W_k(z)  # [batch_size, seq_len, hidden_dim]
        V = self.W_v(z)  # [batch_size, seq_len, hidden_dim]

        # Compute attention scores
        scores = torch.bmm(Q, K.transpose(1, 2))  # [batch_size, seq_len, seq_len]

        mask = causal_mask(seq_len)  # [seq_len, seq_len]
        mask = mask.unsqueeze(0).expand(batch_size, -1, -1)  # [batch_size, seq_len, seq_len]
        scores = scores * mask  # masking with hadamard product for future positions

        # Multiply scores by V to get the output
        output = torch.bmm(scores, V)  # [batch_size, seq_len, hidden_dim]

        return output

class LanguageModel(nn.Module):
    """
    Simple language model for next token prediction, with an embedding layer,
    linear self-attention layer, and output projection.
    """
    def __init__(self, vocab_size, hidden_dim):
        super(LanguageModel, self).__init__()
        self.embedding = nn.Embedding(vocab_size, hidden_dim)
        self.linear_self_attn = LinearSelfAttention(hidden_dim)
        self.fc_out = nn.Linear(hidden_dim, vocab_size)

    def forward(self, z):
        """
        Input z: a batch of sequence with token index 
        [batch_size, seq_len]
        Returns:
        logits: [batch_size, seq_len, vocab_size]
        """
        # Get token embedding
        emb = self.embedding(z)  # [batch_size, seq_len, hidden_dim]

        # Apply masked linear self-attention
        attn_output = self.linear_self_attn(emb)  # [batch_size, seq_len, hidden_dim]

        # Output projection to vocabulary size
        logits = self.fc_out(attn_output)  # [batch_size, seq_len, vocab_size]

        return logits
\end{lstlisting}

\newpage
\subsection{PyTorch example of polynomial nets}
\label{ssec:Had_prod_implementations_pinet}

\rebuttal{Below, we paste two instructional implementations of \pinet s~\cite{chrysos2019polygan}, i.e., the CCP and the NCP models, respectively. Notice that the NCP model implements \cref{eq:prodpoly_model2_simplified}. Sequentially, we paste an instructional implementation of the Poly-SA (cf. \cref{eqn:fast_nl}) for vision applications and MONet from \cite{chen2024multilinear}.}

\begin{lstlisting}[language=python]
from torch import nn

class Pinet(nn.Module):
    def __init__(self, hidden_size=16, image_size=28, channels_in=1, n_degree=4, bias=True, n_classes=10):
        """
        A polynomial-based network (Pi-net) adapted from the CCP model of \pi-nets for image classification.
        Args:
            hidden_size (int): Dimensionality of the hidden representation.
            image_size (int): Height and width of the input images (assumes square images).
            channels_in (int): Number of input channels.
            n_degree (int): Maximum degree of the polynomial expansions.
            bias (bool): Whether to include bias in the linear layers.
            n_classes (int): Number of output classes for the classification task.
        """
        super(Pinet, self).__init__()
        self.image_size = image_size
        self.channels_in = channels_in
        self.total_image_size = self.image_size * self.image_size * channels_in
        self.hidden_size = hidden_size
        self.n_classes = n_classes
        self.n_degree = n_degree
        for i in range(1, self.n_degree + 1):
            setattr(self, 'U{}'.format(i), nn.Linear(self.total_image_size, self.hidden_size, bias=bias))
        self.fc_out = nn.Linear(self.hidden_size, self.n_classes, bias=True)

    def forward(self, z):
        """
        Inputs:
            z: Images of shape [batch_size, channels_in, image_size, image_size].
        Returns:
            torch.Tensor: Logits of shape [batch_size, n_classes].
        """
        h = z.view(-1, self.total_image_size)
        out = self.U1(h)
        for i in range(2, self.n_degree + 1):
            out = getattr(self, 'U{}'.format(i))(h) * out + out
        out = self.fc_out(out)
        return out
\end{lstlisting}

\begin{lstlisting}[language=python]
from torch import nn

class Pinet(nn.Module):
    def __init__(self, hidden_size=16, image_size=28, channels_in=1, n_degree=4, bias=True, n_classes=10):
        """
        A polynomial-based network (Pi-net) adapted from the NCP model of \pi-nets for image classification.
        Args:
            hidden_size (int): Dimensionality of the hidden representation.
            image_size (int): Height and width of the input images (assumes square images).
            channels_in (int): Number of input channels.
            n_degree (int): Maximum degree of the polynomial expansions.
            bias (bool): Whether to include bias in the linear layers.
            n_classes (int): Number of output classes for the classification task.
        """
        super(Pinet, self).__init__()
        self.image_size = image_size
        self.channels_in = channels_in
        self.total_image_size = self.image_size * self.image_size * channels_in
        self.hidden_size = hidden_size
        self.n_classes = n_classes
        self.n_degree = n_degree
        for i in range(1, self.n_degree + 1):
            setattr(self, 'A{}'.format(i), nn.Linear(self.total_image_size, self.hidden_size, bias=bias))
            if i > 1:
                setattr(self, 'S{}'.format(i), nn.Linear(self.hidden_size, self.hidden_size, bias=True))
        self.fc_out = nn.Linear(self.hidden_size, self.n_classes, bias=True)

    def forward(self, z):
        """
        Inputs:
            z: Images of shape [batch_size, channels_in, image_size, image_size].
        Returns:
            torch.Tensor: Logits of shape [batch_size, n_classes].
        """
        h = z.view(-1, self.total_image_size)
        out = self.A1(h)
        for i in range(2, self.n_degree + 1):
            out = getattr(self, 'A{}'.format(i))(h) * getattr(self, 'S{}'.format(i))(out)
        out = self.fc_out(out)
        return out
\end{lstlisting}

\begin{lstlisting}[language=python]
import torch
import torch.nn as nn

class PolySA(nn.Module):
    def __init__(self, channels):
        """
        Implementation of Poly-SA for vision applications. The input are the channels.
        """
        super().__init__()
        self.C = channels
        
        # Learnable parameters W1, W2, W3 (C x C matrices)
        self.W1 = nn.Parameter(torch.Tensor(self.C, self.C))
        self.W2 = nn.Parameter(torch.Tensor(self.C, self.C))
        self.W3 = nn.Parameter(torch.Tensor(self.C, self.C))
        
        # Learnable scalars alpha and beta
        self.alpha = nn.Parameter(torch.Tensor(1))
        self.beta = nn.Parameter(torch.Tensor(1))
        
    
    def phi(self, x):
        # Input x shape: (batch_size, N, C)
        # Average pooling over N (dim=1), then expand back
        avg = torch.mean(x, dim=1, keepdim=True)  # (B, 1, C)
        return avg.expand(-1, x.size(1), -1)      # (B, N, C)
    
    def forward(self, x):
        # Input x shape: (batch_size, N, C)
        # Compute XW1 and XW2
        xw1 = torch.matmul(x, self.W1)  # (B, N, C)
        xw2 = torch.matmul(x, self.W2)  # (B, N, C)
        
        # Hadamard product and apply phi
        hadamard = xw1 * xw2             # (B, N, C)
        phi_out = self.phi(hadamard)     # (B, N, C)
        
        # Hadamard product with input x; then multiplication with W3
        Y_poly = torch.matmul(phi_out * x, self.W3)  # (B, N, C)
        
        # Combine with learnable scalars (skip connections)
        Z = self.alpha * x + self.beta * Y_poly
        return Z

\end{lstlisting}

\begin{lstlisting}[language=python]
class Spatial_Shift(nn.Module):
    def __init__(self):
        super().__init__()
    
    def forward(self, x):
        b,w,h,c = x.size()
        x[:,1:,:,:c//4] = x[:,:w-1,:,:c//4]
        x[:,:w-1,:,c//4:c//2] = x[:,1:,:,c//4:c//2]
        x[:,:,1:,c//2:c*3//4] = x[:,:,:h-1,c//2:c*3//4]
        x[:,:,:h-1,3*c//4:] = x[:,:,1:,3*c//4:]
        return x

class PolyMlp(nn.Module):
    """As used in MONet"""
    def __init__(self, in_features, hidden_features=None, out_features=None,
            bias=True, drop=0., use_conv=False, n_degree=2, use_alpha=True, norm_layer=partial(nn.LayerNorm, eps=1e-6), use_spatial=False):
        super().__init__()
        self.in_features = in_features
        self.out_features = out_features or in_features
        self.hidden_features = hidden_features or in_features
        self.use_alpha = use_alpha
        self.use_spatial = use_spatial
        bias = to_2tuple(bias)
        drop_probs = to_2tuple(drop)
        linear_layer = partial(nn.Conv2d, kernel_size=1) if use_conv else nn.Linear
        self.norm1 =norm_layer(self.hidden_features)
        self.norm3 =norm_layer(self.hidden_features)

        self.n_degree = n_degree
        self.hidden_features = hidden_features
        self.U1 = linear_layer(self.in_features, self.hidden_features, bias=bias)
        self.U2 = linear_layer(self.in_features, self.hidden_features//8, bias=bias)
        self.U3 = linear_layer(self.hidden_features//8, self.hidden_features, bias=bias)
        self.C = linear_layer(self.hidden_features, self.out_features, bias=True) 
        self.drop2 = nn.Dropout(drop_probs[0])
        
        if self.use_spatial:
            self.spatial_shift = Spatial_Shift()
        if self.use_alpha:
            self.alpha = nn.Parameter(torch.ones(1))
        self.init_weights()
    
    def init_weights(self):
        nn.init.kaiming_normal_(self.U1.weight)
        nn.init.kaiming_normal_(self.U2.weight)
        nn.init.kaiming_normal_(self.U3.weight)
        nn.init.ones_(self.U1.bias)
        nn.init.ones_(self.U2.bias)
        nn.init.ones_(self.U3.bias)
            
    def forward(self, x):  #
        if self.use_spatial:               
            out1 = self.U1(x)             
            out2 = self.U2(x)       
            out1 = self.spatial_shift(out1)
            out2 = self.spatial_shift(out2)
            out2 = self.U3(out2) 
            out1 = self.norm1(out1)
            out2 = self.norm3(out2)
            out_so = out1 * out2
        else:
            out1 = self.U1(x)          
            out2 = self.U2(x)
            out2 = self.U3(out2)
            out1 = self.norm1(out1)
            out2 = self.norm3(out2)
            out_so = out1 * out2
        if self.use_alpha:
            out1 = out1 + self.alpha * out_so
            del out_so
        else:
            out1 = out1 + out_so
            del out_so
        out1 = self.C(out1)
        return out1
    

class PolyBlock(nn.Module):
    def __init__(self, embed_dim, expansion_factor = 3, mlp_layer = PolyMlp,
            norm_layer=partial(nn.LayerNorm, eps=1e-6), drop=0., drop_path=0., n_degree = 2):
        super().__init__()
        self.embed_dim = embed_dim
        self.expansion_factor = expansion_factor
        self.norm = norm_layer(self.embed_dim)
        self.drop_path = DropPath(drop_path) if drop_path > 0. else nn.Identity()
        self.mlp1 = PolyMlp(self.embed_dim, self.embed_dim, self.embed_dim, drop=drop, use_spatial=True)
        self.mlp2= PolyMlp(self.embed_dim, self.embed_dim*self.expansion_factor, self.embed_dim, drop=drop, use_spatial=False)
    
    def forward(self, x):
        z = self.norm(x)
        z = self.mlp1(z)  
        x = x + self.drop_path(z)
        z = self.norm(x)
        z = self.mlp2(z)
        x = x + self.drop_path(z)
        return x


class Downsample(nn.Module):
    """ Downsample transition stage   design for pyramid structure
    """
    def __init__(self, in_embed_dim, out_embed_dim, patch_size):
        super().__init__()
        assert patch_size == 2, patch_size
        self.proj = nn.Conv2d(in_embed_dim, out_embed_dim, kernel_size=(3, 3), stride=(2, 2), padding=1)

    def forward(self, x):
        # x = rearrange(x, 'b c h w -> b h w c')
        x = self.proj(x)
        # x = rearrange(x, 'b h w c -> b c h w')
        # x = x.permute(0, 3, 1, 2)
        # x = self.proj(x)  # B, C, H, W
        # x = x.permute(0, 2, 3, 1)
        return x
    


class MONet(nn.Module):
    def __init__(
        self,
        image_size=224,
        num_classes=1000,
        in_chans=3,
        patch_size= 2,
        mlp_ratio = [0.5, 4.0],
        block_layer =basic_blocks,
        mlp_layer = PolyMlp,
        norm_layer=partial(nn.LayerNorm, eps=1e-6),
        drop_rate=0.,
        drop_path_rate=0.,
        nlhb=False,
        global_pool='avg',
        transitions = None,
        embed_dim=[192, 384],
        layers = None,
        expansion_factor = [3, 3],
        feature_fusion_layer = None,
        use_multi_level = False,
    ):
        # self, layers, img_size=224, patch_size=4, in_chans=3, num_classes=1000,
        # embed_dims=None, transitions=None, segment_dim=None, mlp_ratios=None,  drop_rate=0., attn_drop_rate=0., drop_path_rate=0.,
        # norm_layer=nn.LayerNorm, mlp_fn=CycleMLP, fork_feat=False
        self.num_classes = num_classes
        self.image_size = image_size
        self.global_pool = global_pool
        self.num_features = self.embed_dim = embed_dim[-1]  # num_features for consistency with other models
        self.use_multi_level = use_multi_level
        self.grad_checkpointing = False
        self.layers = layers
        self.embed_dim = embed_dim
        image_size = pair(self.image_size)
        oldps = [1, 1]
        for ps in patch_size:
            ps = pair(ps)
            oldps[0] = oldps[0] * ps[0]
            oldps[1] = oldps[1] * ps[1]
        super().__init__()
    
        self.fs = nn.Conv2d(in_chans, embed_dim[0], kernel_size=patch_size[0], stride=patch_size[0])
        self.fs2 = nn.Conv2d(embed_dim[0], embed_dim[0], kernel_size=2, stride=2)
        network = []
        assert len(layers) == len(embed_dim) == len(expansion_factor)
        for i in range(len(layers)):
            stage = block_layer(i, self.layers, embed_dim[i], expansion_factor[i], dropout = drop_rate,drop_path=drop_path_rate,norm_layer=norm_layer)
            network.append(stage)
            if i >= len(self.layers)-1:
                break
            if transitions[i] or embed_dim[i] != embed_dim[i+1]:
                patch_size = 2 if transitions[i] else 1
                network.append(Downsample(embed_dim[i], embed_dim[i+1], patch_size))
        self.network = nn.Sequential(*network)
        self.head = nn.Sequential(
            Reduce('b c h w -> b c', 'mean'),
            nn.Linear(embed_dim[-1], self.num_classes)
        )
        self.init_weights(nlhb=nlhb)
        
    def forward(self, x):
        x1 = self.fs(x)
        x1 = self.fs2(x1)
        if self.use_multi_level:
            x2 = self.fs3(x)
            x1 = x1 + self.alpha1 * x2
        embedding = self.network(x1)
        out = self.head(embedding)
        return out
    
    def forward_features(self, x):
        x1 = self.fs(x)
        x1 = self.fs2(x1)
        if self.use_multi_level:
            x2 = self.fs3(x)
            x1 = x1 + self.alpha1 * x2
        embedding = self.network(x1)
        return embedding
\end{lstlisting}

\rebuttal{Alongside these implementations, the authors have also developed introductory code for some of the networks discussed in this survey. This code is already publicly accessible \href{https://github.com/polynomial-nets/tutorial-2022-intro-polynomial-nets/tree/master}{here}. The link includes PyTorch, Tensorflow, Jax and Keras implementations.}

\section{Numerical efficiency of Hadamard product}

\rebuttal{In \cref{sec:elementwise_pr_efficient_op}, we discuss the computational efficiency of the Hadamard product. This efficiency has been empirically validated in numerous studies, including \cite{babiloni2023linear}. As shown in \cref{fig:hadamard_product_computation_efficiency_polynl,fig:hadamard_product_computation_efficiency_polysa}, these studies demonstrate the efficiency in vision and text applications compared to networks that do not utilize the Hadamard product.}

\begin{figure*}[t]
    \hspace*{-0.7cm}
      \subfloat[\centering]{\label{fig:big_c}\includegraphics[trim=0 0 10 0,clip,width=.356\linewidth, ]{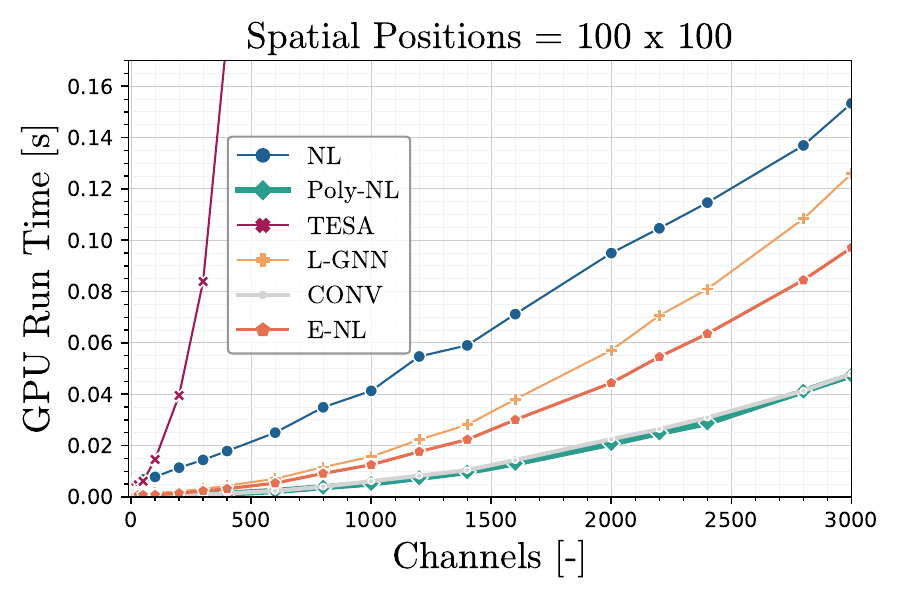}}   \subfloat[]{\label{fig:big_s}\includegraphics[trim=0 0 10 0,clip,width=.356\linewidth, ]{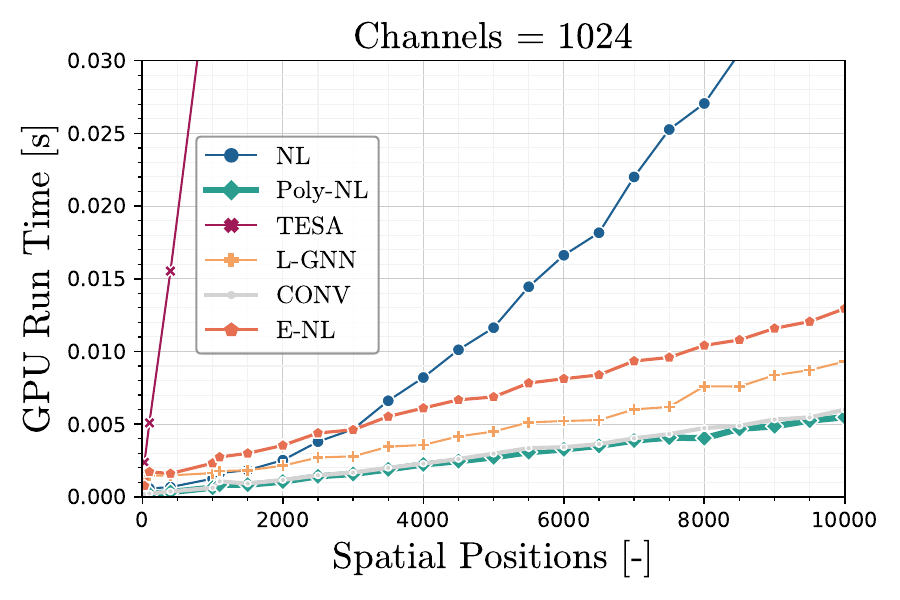}} 
      \subfloat[\centering  ]{\label{fig:small_s}\includegraphics[trim=0 0 10 0,clip,width=.356\linewidth, ]{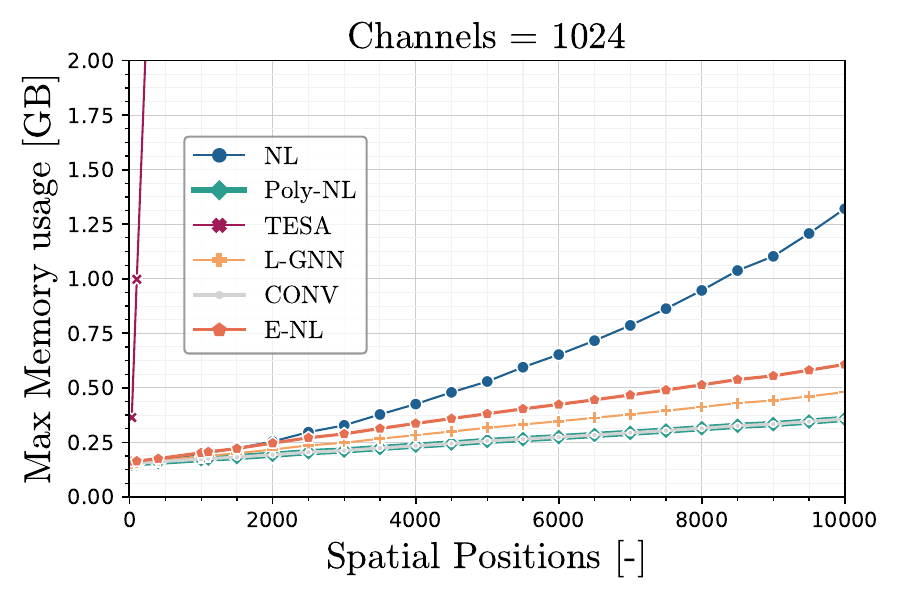}}   \vspace{-0.25cm}
      \caption{\rebuttal{\textbf{Runtime and Peak memory consumption} performance comparison in vision between Poly-NL, which implements (a variant of) \cref{eqn:fast_nl}, and other non-local methods executed on a RTX2080 GPU.
      The network utilizing the Hadamard product exhibits lower computational overhead than competing methods, which is of importance with an increasing number of spatial positions or channels. The figure is reproduced from \cite{babiloni2023linear}. \vspace{-0.1cm}}}
      \label{fig:hadamard_product_computation_efficiency_polynl}
\end{figure*}

\begin{figure*}[t]
    \hspace*{-0.7cm}
      \subfloat[\centering]{\label{fig:transf_run_fc}\includegraphics[trim=0 0 10 0,clip,width=.356\linewidth, ]{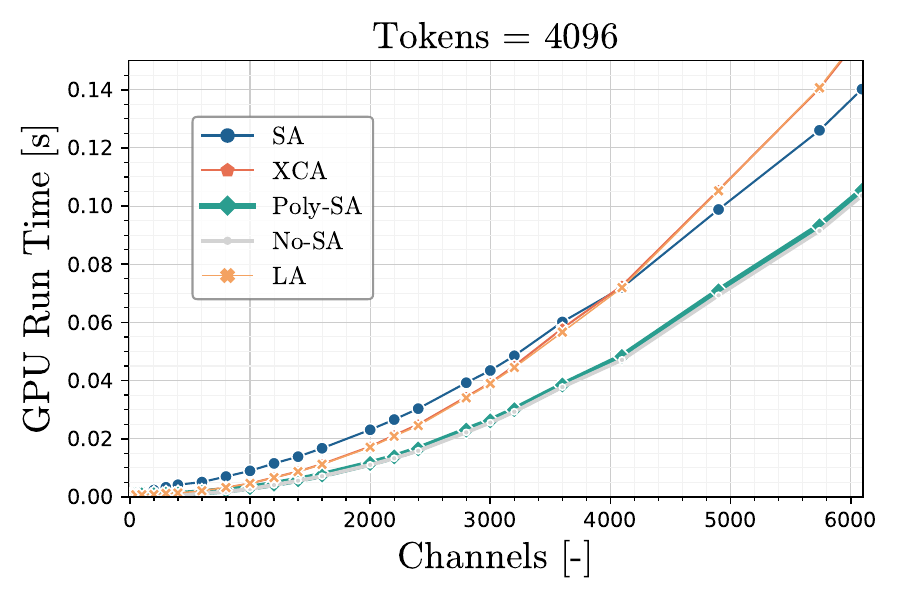}}   \subfloat[\centering]{\label{fig:transf_run_fs}\includegraphics[trim=0 0 10 0,clip,width=.356\linewidth, ]{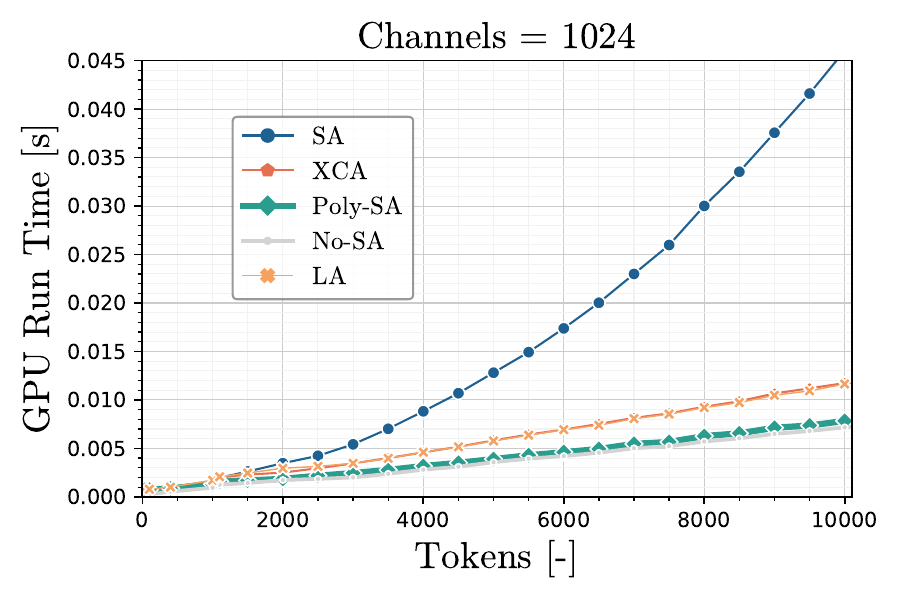}} 
      \subfloat[\centering  ]{\label{fig:transf_FLOPS_fs}\includegraphics[trim=0 0 10 0,clip,width=.356\linewidth, ]{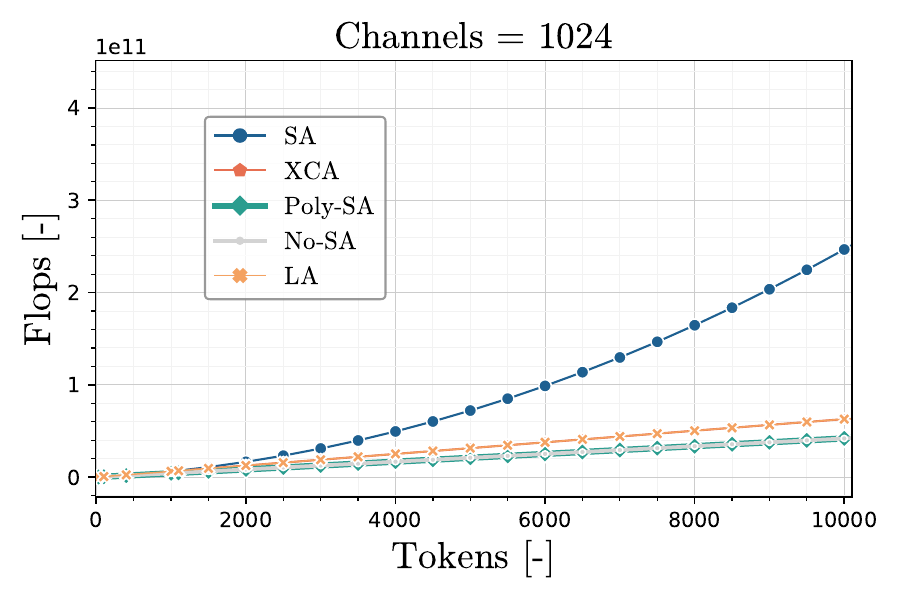}}   \vspace{-0.25cm}
      \caption{ \rebuttal{\textbf{Runtime and Flops} comparison in text domain between Poly-SA, which implements (a variant of) \cref{eqn:fast_nl}, and two other self-attention method executed on a RTX2080 GPU. 
      The network utilizing the Hadamard product exhibits lower computational overhead than competing methods, with a complexity comparable to a linear layer using no attention mechanism. The figure is reproduced from \cite{babiloni2023linear}.\vspace{-0.1cm}}}
      \label{fig:hadamard_product_computation_efficiency_polysa}
\end{figure*}

\section{\rebuttal{Hadamard product in large language models}}
\rebuttal{
In this section, we review the use of the Hadamard product in large language models (LLMs). Related literature is summarized in \cref{tab:hadamard_product_indicative_work_table}.}

\rebuttal{Within the category of adaptive modulation, the Hadamard product appears in LLMs in two main ways. Firstly, it is used in masked language modeling, as discussed in \cref{eq:lin_attention} of the main body. During LLM pre-training, the objective is to predict the next token based on all preceding tokens. A masking mechanism enables this prediction to occur in a single forward pass, while ensuring that a query cannot access keys corresponding to future tokens. An indicative PyTorch implementation is provided in \cref{ssec:Had_prod_implementations_linear_attention_masked}.
Secondly, the Hadamard product functions as a gating mechanism in the core architecture blocks of LLMs, as detailed in \cref{ssec:elementwise_pr_rnn}. For example, recent models such as xLSTM~\cite{hochreiter1997long}, Mamba~\cite{gu2024mamba,dao2024transformers}, and others~\cite{qin2024hgrn,yang2024gated} have adopted the Hadamard product in this role.
}

\rebuttal{
Beyond adaptive modulation, the Hadamard product is also employed in parameter-efficient fine-tuning methods for LLMs~\cite{hyeonwoo2022fedpara,anonymous2025hira,wang2024mlae}. Given a pretrained weight matrix $\bm{W}^0$, standard LoRA assumes that the fine-tuned weight $\bm{W}^\star$ satisfies
$$
\bm{W}^\star = \bm{W}^0 + \Delta \bm{W} = \bm{W}^0 + \bm{A} \bm{B},
$$
where $\bm{A} \bm{B}$ represents a low-rank adaptation. A recent method, HIRA~\cite{anonymous2025hira}, modifies this update rule to retain high-rank updates and enhance model capacity:
$$
\bm{W}^\star = \bm{W}^0 + \bm{W}^0 \hp ( \bm{A} \bm{B} ).
$$
In addition to HIRA, several other LoRA variants incorporating the Hadamard product have also been proposed~\cite{hyeonwoo2022fedpara,wang2024mlae}.
}

\end{document}